\documentclass{article}

\PassOptionsToPackage{numbers, compress}{natbib}
\usepackage[preprint]{neurips_2022}

\usepackage[utf8]{inputenc} % allow utf-8 input
\usepackage[T1]{fontenc}    % use 8-bit T1 fonts
\usepackage{hyperref}       % hyperlinks
\usepackage{url}            % simple URL typesetting
\usepackage{booktabs}       % professional-quality tables
\usepackage{amsfonts}       % blackboard math symbols
\usepackage{nicefrac}       % compact symbols for 1/2, etc.
\usepackage{microtype}      % microtypography
\usepackage{xcolor}         % colors

\usepackage{natbib} % has a nice set of citation styles and commands
    \renewcommand{\bibsection}{\subsubsection*{References}}
    
\usepackage{microtype}
\usepackage{graphicx}
\usepackage{booktabs} % for professional tables
\usepackage{comment}

\usepackage{amsmath}
\usepackage{amssymb}
\usepackage{amsthm}

\usepackage{amsmath,amsfonts,bm}

\def\eqref#1{equation~\ref{#1}}
\def\1{\bm{1}}

\DeclareMathAlphabet{\mathsfit}{\encodingdefault}{\sfdefault}{m}{sl}
\SetMathAlphabet{\mathsfit}{bold}{\encodingdefault}{\sfdefault}{bx}{n}

\DeclareMathOperator*{\argmax}{arg\,max}
\DeclareMathOperator*{\argmin}{arg\,min}

\usepackage{url}
\usepackage{comment}

\usepackage[utf8]{inputenc} % allow utf-8 input
\usepackage[T1]{fontenc}    % use 8-bit T1 fonts
\usepackage{url}            % simple URL typesetting
\usepackage{amsfonts}       % blackboard math symbols
\usepackage{nicefrac}       % compact symbols for 1/2, etc.
\usepackage{xcolor}         % colors
\usepackage{arydshln}
\usepackage{pdfpages}

\usepackage{caption}
\usepackage{subcaption}
\usepackage{listings}
\usepackage{amsmath}
\usepackage{colortbl} %use in the preamble
\usepackage{multirow}

\usepackage{mathtools,xparse}
\usepackage{nicefrac}       % compact symbols for 1/2, etc.
\usepackage{wrapfig}

\usepackage{algpseudocode, algorithm}
\makeatletter
\renewcommand{\ALG@beginalgorithmic}{\small}
\makeatother

\theoremstyle{plain}
\newtheorem{theorem}{Theorem}[section]

\theoremstyle{definition}
\newtheorem{definition}[theorem]{Definition}

\theoremstyle{remark}

\usepackage[textsize=tiny]{todonotes}

\newcommand{\edit}[1]{\textcolor{blue}{#1}}

\newcommand{\algname}{BiLAW}

\title{Learning Sample Reweighting for \\ Accuracy and Adversarial Robustness}

\author{%
  Chester Holtz \\
  Computer Science and Engineering\\
  University of California San Diego\\
  La Jolla, CA 92093 \\
  \texttt{chholtz@eng.ucsd.edu}
  \And
  Tsui-Wei Weng \\
  Halicio\v{g}lu Data Science Institute\\
  University of California San Diego\\
  La Jolla, CA 92093 \\
  \texttt{lweng@ucsd.edu}
  \And
  Gal Mishne \\
  Halicio\v{g}lu Data Science Institute\\
  University of California San Diego\\
  La Jolla, CA 92093 \\
  \texttt{gmishne@ucsd.edu}
}

\begin{document}

\maketitle

\begin{abstract}
There has been great interest in enhancing the robustness of neural network classifiers to defend against adversarial perturbations through adversarial training, while balancing the trade-off between robust accuracy and standard accuracy. We propose a novel adversarial training framework that learns to reweight the loss associated with individual training samples based on a notion of class-conditioned margin, with the goal of improving robust generalization. We formulate weighted adversarial training as a bilevel optimization problem with the upper-level problem corresponding to learning a robust classifier, and the lower-level problem corresponding to learning a parametric function that maps from a sample's \textit{multi-class margin} to an importance weight. Extensive experiments demonstrate that our approach consistently improves both clean and robust accuracy compared to related methods and state-of-the-art baselines. 
\end{abstract}

\section{Introduction}
%\lily{I feel most part of the bi-level optimization and some part of the adv overfitting should be put into related work section? As it's not so smooth after reading adv overfitting then jump to bi-level optimization. In the intro, we could have 1 paragraph talking about adv overfitting as motivation, and 1 paragraph about what we are proposing and why solving via bi-level optimization is a good idea. Then follow by the current contribution paragraph}

\label{introduction}

\looseness=-1
While neural networks have been extremely successful in tasks such as image classification and speech recognition, recent work~\citep{szegedy2014intriguing, goodfellowadv} has demonstrated that neural network classifiers can be arbitrarily fooled by small, adversarially-chosen perturbations of their input. 
Notably, \citet{SuSinglePixel17} demonstrated that neural network classifiers which can correctly classify ``clean'' images may be vulnerable to \textit{targeted attacks}, e.g., misclassify those same images when only a single pixel is changed.

%\textbf{Adversarial overfitting} 
\looseness=-1
Recent work has shown a common failing among techniques that uniformly encourage robustness. 
In particular, there exists an intrinsic tradeoff between robustness and accuracy~\citep{trades}.
\citet{bao20classcall} investigate this tradeoff  from the perspective of classification-callibrated loss theory. \citet{riceadvoverfitting} empirically showed that during adversarial training networks often irreversibly lose robustness after training for a short time. They dubbed this phenomenon \textit{adversarial overfitting} while proposing early stopping as a remedy. The significance of label noise and memorization in the context of adversarial overfitting was demonstrated by~\citet{sanyal2021benign}\textemdash in particular that poor training samples induce fragility to adversarial perturbations due to the tendency of neural networks to interpolate the training data. Methods based on weight and logit smoothing have been proposed as an alternative to early stopping~\citep{chen2021robustsmooth} as well as techniques for dataset augmentation~\cite{rebuffi2021data, gowal2021} and local smoothing~\cite{yang2020closerlook,localLipschitz}.

\looseness=-1
In a different approach to addressing adversarial overfitting, Geometry-Aware Instance Reweighted Adversarial Training (GAIRAT;~\cite{zhang2021geometryaware}), Weighted Margin-aware Minimax Risk (WMMR;~\cite{zeng2020adversarial}), and Margin-Aware Instance reweighting Learning (MAIL;~\cite{wang2021probabalisticmarginat}) control the influence of training examples via importance or loss weighting. 
Intuitively, the samples assigned a low weight correspond to samples on which the classifier is already sufficiently robust. Generally, these methods are well-motivated\textemdash e.g. by \cite{xu2021understanding} who conclude that a good set of weights (large (small) weights for samples close (far) to the decision boundary) are tied to generalization. However, existing methods rely on approximations of the margin and employ heuristic weighting schemes that rely on careful choices of hyperparameters. 

\looseness=-1
Building upon these observations, we present BiLAW (Bilevel Learnable Adversarial reWeighting), an approach that explicitly learns a parametric function (e.g. represented by a small feed-forward network) that assigns weights to the loss suffered by a classifier, associated with individual training samples.
The sample weights are learned as a function of the classifier \textit{multiclass margins} of samples, according to the weights' effect on \textit{robust generalization}.
%automatically learn to map between the \textit{multiclass margin} of an example to its associated weight according to its effect on robust generalization.
We employ a bi-level optimization formulation~\cite{braken1973bilevel} and leverage a validation set, where the \textit{upper-level} objective corresponds to learning the parameters of a robust classifier, while the \textit{lower-level} objective corresponds to learning a function that predicts sample weights that improve robustness on a validation set. 
% https://openreview.net/pdf?id=sMEpviTLi1h
% https://arxiv.org/pdf/1809.01465.pdf
%\textbf{Meta-learning \& bilevel optimization} 
%In this work, we apply the framework of bilevel optimization to learn a parametric function (e.g. represented by a small feed forward network) that assigns weights to the loss suffered by a classifier associated with individual training samples. 
Our approach alternates between iteratively updating the parametric sample weights and updating the classifier network parameters.
%The parameters of this function are iteratively re-adjusted alongside training the classifier.%so as to assign weights that improve robust generalization. 

\textbf{Contributions}  As far as we know, this is the first work to explore a \textit{learning}-based approach to sample weighting in the context of adversarial training. Prior work~\citep{zhang2021geometryaware, yi2021reweighting, wang2021probabalisticmarginat} only used heuristics to estimate the weight and did not involve any learning components. Our contributions include:
% We note that while sample weighting has been investigated in the context of robust training,
% a method to use a validation set to learn weights to induce robust generalization. 
%
\begin{enumerate}
     \item We propose BiLAW, a new adversarial training method based on learning sample weights as a parametric function mapping from multi-class margins. Our method can be formulated as a bi-level optimization problem that can be solved efficiently thanks to recent advances in meta-learning. 
    %  to weights according to the robust loss suffered by a classifier on a validation set, and formulate this as a bilevel optimization problem.
     %We propose a bilevel optimization formulation to learn a mapping from multi-class margins to weights according to the robust loss suffered by a classifier on a validation set.
    \item We motivate and extend the notion of the robust margin of a classifier at a particular sample to the multi-class setting, and show that %when our weighting function corresponds to a neural network, 
    the magnitude of a sample's learned weight directly corresponds to the vulnerability of the classifier at that sample. 
    % \lily{this contribution is a bit thin, need more description. Are you referring to Fig 3?}
    %\item We evaluate the practical performance of \algname{} on MNIST~\citep{lecun-mnisthandwrittendigit-2010}, F-MNIST~\citep{Xiao17}, CIFAR-10, and CIFAR-100.~\citep{KrizhevskyCIFAR10} and demonstrate that it improves clean accuracy up to $6\%$ and robust test accuracy by up to $5\%$ compared to TRADES and other state-of-the-art  sample reweighting methods on CIFAR-10. %when applied to the TRADES robust loss.
    \item We evaluate the performance of \algname{} on MNIST, %~\citep{lecun-mnisthandwrittendigit-2010}, 
    F-MNIST%~\citep{Xiao17}
    , and CIFAR-10 and demonstrate it significantly improves clean accuracy by up to $6\%$ and robust test accuracy by up to $5\%$ compared to TRADES and other state-of-the-art  sample reweighting methods on CIFAR-10. %when applied to the TRADES robust loss.
\end{enumerate}
%
%This paper is organized as follows. Section~\ref{sec:preliminaries} reviews the notation and background of cost-aware and robust classification. In Section~\ref{sec:method} we describe the learnable sample weighting method. In Section~\ref{sec:experiments}, we provide ablative experiments and demonstrate the efficacy of our framework by comparing clean and robust test performance on MNIST, F-MNIST, CIFAR-10, and CIFAR-100 to adversarial training and two recent, state-of-the-art sample weighting methods. 
%
\section{Preliminaries and Related Work}
\label{sec:preliminaries}

In this section, we briefly present background terminology pertaining to adversarially robust classification, sample reweighting and bilevel optimization.

\paragraph{Notations} %An ReLU network is a neural network such that all nonlinear activations are ReLU functions, where we denote the ReLU activation by $\sigma : \mathbb{R} \to \mathbb{R}$, $\sigma(x) = \max\{0,x\}$. Informally, we define $\sigma : \mathbb{R}^d \to \mathbb{R}^d$ by $\sigma(x) = [\sigma(x_1), \ldots, \sigma(x_d)]$. 
Let $f : \mathbb{R}^d \to [0,1]^k$ be a feedforward ReLU network with $l$ hidden layers and weights $\theta$; for example, $f$ may map from a $d$-dimensional image to a $k$-dimensional vector corresponding to likelihoods for $k$ classes. 

Given a training set of $m$ sample-label pairs $(x_i,y_i)$ drawn from a training data distribution $\mathcal{D}$, we associate a \emph{weight} $w_i$ with each training sample.
Informally, these weights characterize the effect of the sample on the generalization of the network (i.e. samples with large weights promote robust generalization and visa versa).
Given a loss function $\ell : \mathbb{R}^k \times \mathbb{R}^k \to \mathbb{R}$, we denote the empirical \textit{weighted} training loss suffered by a network with parameters $\theta$ on $m$ training samples with weights $w$ to be $\mathcal{L}_\textrm{tr}(\theta, w) = \sum_{i=1}^{m}w_i\ell(y_i, f(x_i; \theta))$ such that $w_i \geq 0$ and $\sum_{i}w_i = 1$. For brevity, we write $\ell_i(\theta) = \ell(y_i, f(x_i; \theta))$. Additionally, if $w$ is left unspecified, $\mathcal{L}$ corresponds to the unweighted mean over empirical losses. Likewise, the \textit{unweighted} validation loss of $n$ samples is denoted $\mathcal{L}_{\textrm{val}}(\theta)$.

%Let $n_l$ be the number of hidden units at layer $l$, the input layer is of size $n_0=d$, and let $W^{(l)}\in \mathbb{R}^{n_{l-1}\times n_l}$ and $b^{(l)} \in \mathbb{R}^{n_l}$ denote the weight matrix and bias vector at layer $l$, respectively.

\subsection{Robust classification and adversarial overfitting}
Consider the network $f:\mathbb{R}^d \to \mathbb{R}^k$, where the input is $d$-dimensional and the output is a $k$-dimensional vector of likelihoods, with $j$-th entry corresponding to the likelihood the image belongs to the $j$-th class. The associated classification is then $c(x; \theta) = \argmax_{j\in[1,k]} f_{j}(x; \theta)$. In adversarial machine learning, we are not just concerned that the classification be correct, but we also want to be robust against adversarial examples, i.e. small perturbations to the input which may change the classification to an incorrect class. We define the notion of $\epsilon$-robustness below:
\begin{definition}[$\epsilon$-robust]
$f$ parameterized by $\theta$ is called $\epsilon$-robust with respect to norm $p$ at $x$ if the classification is consistent for a small ball of radius $\epsilon$ around $x$:
\begin{equation}
    \label{eq:erobust}
    c(x+\delta; \theta) = c(x; \theta), \forall \delta : ||\delta||_p \leq \epsilon.
\end{equation}
\end{definition}
Note that the $\epsilon$-robustness of $f$ at $x$ is intimately related to the uniform Lipschitz smoothness of $f$ around $x$. Recall that a function $f$ has finite Lipschitz constant $L > 0$ with respect to norm $||\cdot||$, if
\begin{equation}
\label{eq:lipschitz}
    \exists L \geq 0 \text{ s.t. } |f(x) - f(x')| \leq L \cdot ||x - x'||, \forall x, x' \in X.
\end{equation}
An immediate consequence of Eq.~\ref{eq:erobust} and Eq.~\ref{eq:lipschitz} is that if $f$ is uniformly $L$-Lipschitz, then $f$ is $\epsilon$-robust at $x$ with $\epsilon = \frac{1}{2L}(P_a - P_b)$ where $P_a$ is the likelihood of the most likely outcome, and $P_b$ is the likelihood of the second most likely outcome \citep{SalmanSmooth4}. The piecewise linearity of ReLU networks facilitates the extension of this consequence to the \textit{locally Lipschitz} regime~\cite{yang2020closerlook, localLipschitz}. $L$ corresponds to the norm of the affine map characterized by $f$ conditioned on input $x$. These properties were previously  \citep{zeng2020adversarial, yi2021reweighting, wang2021probabalisticmarginat} used to characterize the robustness of a network at a sample (and the weight associated with the sample). % according to its \textit{margin} \citep{zeng2020adversarial, yi2021reweighting, wang2021probabalisticmarginat}. 

The minimal $\ell_p$-norm perturbation $\delta_p^*$ required to switch an sample's label is given by the solution to the following optimization problem:
$$
\delta^*_p = \arg\min ||\delta||_p \quad \textrm{s.t.}\quad c(x; \theta) \neq c(x+\delta; \theta).
$$
A significant amount of existing work relies on a first-order approximations and H\"{o}lder's inequality to recover $\delta^*$,
%
\iffalse

$$
r^*_p = \arg\min ||r||_p \quad \textrm{s.t.}\quad \langle\nabla f, r\rangle \leq -\delta
$$
where $\delta$ is the error associated with the first-order approximation (e.g. $f(x+r) = f(x) + \langle\nabla f, r\rangle + \delta$).
%
\begin{align*}
r^* &\approx \frac{\delta}{||\nabla f||_q}\partial(||\nabla f||_q) \\
&= \frac{\delta}{||\nabla f||_q}(\frac{|\nabla f|^{q-1}\odot\textrm{sign}(\nabla f)|}{||\nabla f||_q^{q-1}})
\end{align*}
%
\fi
%
justifying the popularity of inducing robustness by controlling global and local Lipschitz constants.
More concretely, given a $\ell_p$ norm and radius $\epsilon$, a typical goal of robust machine learning is to learn classifiers that minimize the \textit{robust loss} on a training dataset:
$$
\min_\theta \mathbb{E}_{(x,y)\sim \mathcal{D}}\left[\max_{ ||\delta||_p \leq \epsilon}\ell(y,f(x+\delta; \theta))\right].
$$
For brevity we will denote the robust analogue of a loss $\mathcal{L}$ as $\hat{\mathcal{L}}$ (likewise, the pointwise loss $\ell$ as $\hat{\ell}$), indicating this is the robust counterpart of $\mathcal{L}$, differentiated by the ``inner'' maximization problem.
\begin{figure}
    \includegraphics[width=0.99\textwidth]{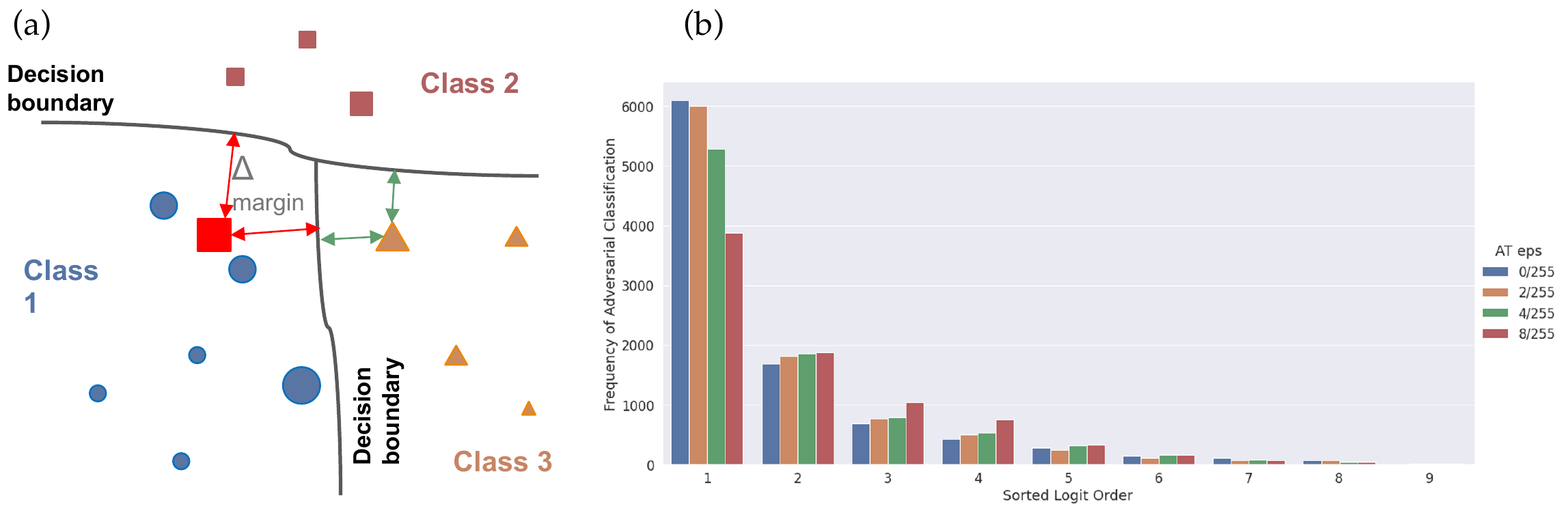}
   % \vspace{-1cm}
\caption{(\textbf{a}) Diagram of multiclass margin. Larger samples denote samples that should be assigned large weight, e.g., are misclassified or close to the decision boundary. Green (red) arrows denote entries in the multiclass margin vector for a correctly (incorrectly) classified sampled. (\textbf{b}) Sorted logit order and frequency of adversarial classification. Number of instances where the prediction of an adversarial sample corresponds to its $i$-th largest logit in CIFAR-10 (ignoring the $0$-th logit/samples where the prediction does not change). Colors represent the perturbation budget used during adversarial training (i.e. degrees of robustness). Perturbations are computed using $\ell_\infty$-PGD with 10 iterations and a budget of $0.031$.}%\lily{Figure can be made larger, currently too small}
\label{fig:multiclassmargin}
\vspace{-0.5cm}
\end{figure}
\subsection{Margin-aware Reweighting}
%Cost-sensitive learning}

%In this work, we exploit an uncorrupted validation set to jointly learn weights on training samples and classifiers which minimize the associated weighted robust error. Intuitively, the samples with high weight should improve robust generalization\textemdash this is quantified by the robust error evaluated on a held-out validation set. 

%We propose to learn to weight samples according to their influence on the robust generalization of the classifier. Let $\hat{\mathcal{L}}_{\textrm{tr}}(\theta_t, w_t) = \sum_{i=1}^{m_{\textrm{tr}}}w_{t,i}\hat{\ell}_i(\theta_t)$ be the \textit{weighted} robust training loss with respect to parameters $\theta_t$ at time $t$ with sample weights $w_{t}$, where $w_{t,i} = f_\mu(\ell_{\textrm{tr},i}(\tilde{\theta}))$. 

In the framework of cost-sensitive learning, weights are assigned to the loss associated with individual samples and the goal is to minimize the empirical \textit{weighted training loss}:
$$
\mathcal{L}_{\textrm{tr}}(\theta, w) := \sum_{i=1}^{m}w_{i}\ell_i(\theta).
$$
Previous work in margin-aware adversarial training \citep{zhang2020fat, zhang2021geometryaware, zeng2020adversarial, Balaji2019InstanceAA, Ding2020MMA} typically substitutes the robust loss $\hat{\mathcal{L}}_\textrm{tr}$ for $\mathcal{L}_\textrm{tr}$ and largely focuses on designing heuristic functions of various notions of margin to use for the sample weight $w_i$.
%Having defined the margin, the weight associated with a particular sample then corresponds to a carefully designed function of the margin of a classifier at that sample.
%As mentioned preivously, prior work in cost-sensitive learning for adversarial robustness focuses on designing heuristic functions of different notions of margin.

%For example, In GAIRAT~\cite{zhang2021geometryaware, zhang2020fat} define the margin as the least number of PGD steps for a sample to be misclassified.
For example, in GAIRAT~\citep{zhang2021geometryaware, zhang2020fat, Ding2020MMA}, the margin is defined as the least number of PGD steps, denoted $\kappa$, that leads the classifier to make an incorrect prediction. The sample's weight is computed as 
% For example, \cite{zhang2021geometryaware} choose the 
$\omega_{\textrm{GAIRAT}}(x_i) = \frac{1}{2}(1+\tanh(\lambda + 5(1-2\kappa/K)))$ with hyperparameters $K$ and $\lambda$.
A small $\kappa$ indicates that the sample lies close to the decision boundary. Larger $\kappa$ values imply that associated samples lie far from the decision boundary, and are therefore more robust, requiring smaller weights. However, due to the non-linearity of the loss-surface in practice, PGD-based attacks with finite iterations may  suffer from the same issues that plague standard iterative first-order methods in non-convex settings. In other words, $\kappa$ is heavily dependent on the optimization path taken by PGD. 
This is demonstrated by GAIRAT's vulnerability
%substandard performance 
to sophisticated attacks, e.g. AutoAttack~\citep{croce2020reliable}.

\citet{zhang2020fat} define the margin as the difference between the loss of a network suffered at a clean sample and its adversarial variant. \citet{zeng2020adversarial, wang2021probabalisticmarginat, Balaji2019InstanceAA} propose a definition of margin corresponding to taking differences between logits, as follows.
\begin{definition}[\citet{zeng2020adversarial, wang2021probabalisticmarginat}]\label{def:cm} The margin of a classifier $f$ on sample $(x
, y)$ is the difference between the confidence of $f$ in the true label $y$ and the maximal probability of an incorrect label $t$, $\textrm{margin}(x , y; \theta) = p(f(x; \theta) = y) - \max_{t\neq y} p(f(x; \theta) = t)$.
\end{definition}

Given this definition, \citet{zeng2020adversarial, wang2021probabalisticmarginat} propose to use exponential (WMMR) and sigmoidal (MAIL) functions respectively: $\omega_{\textrm{WMMR}}(x_i) = \exp(-\alpha m)$ with parameter $\alpha$, and $\omega_{\textrm{MAIL}}(x_i) = \textrm{sigmoid}(-\gamma(m-\beta))$ with parameters $\gamma$ and $\beta$.
WMMR and MAIL %
%utilize a notion of margin defined using the output logits. These methods 
rely on the local linearity of ReLU networks and that for samples near the margin, the relative scale of predicted class-likelihoods directly corresponds to the distance to the decision boundary. However, similarly to GAIRAT's $\kappa$, even for samples very close to the decision boundary, simple functions of the difference between class likelihoods may not necessarily correspond to the true distance to the decision boundary. In contrast, we propose a more fine-grained notion of margin, the \textit{multi-class margin}, and a method to learn a mapping between the margin at a sample and its associated weight, rather than use a predefined heuristic function.

Previous work has explored theoretical notions of a \textit{multi-class margin}. For example, \citet{Zou2005TheMV} defined the \textit{margin vector} in the context of boosting as a proxy for a vector of conditional class probabilities. %They showed that when used in conjunction with an \textit{admissible loss}, the class with the largest conditional probability has the largest margin. 
However, this notion of margin is unaware of the true class of a sample. In contrast, the multi-class margin proposed by \citet{saberianmulticlassboosting19, cortesmulticlasskernel13} are both closely related to~\citet{wang2021probabalisticmarginat, zeng2020adversarial}, i.e. defined as the minimal distance between an arbitrary predicted logit and the logit of the true class.

%
\iffalse
\begin{figure}
    \centering
    \begin{subfigure}[b]{0.54\linewidth}
    \centering
    \includegraphics[width=0.8\textwidth]{figures/margin_diagram.png}
    \caption{Caption}
    \label{fig:margin_diagram}
    \end{subfigure}
    \begin{subfigure}[b]{0.45\linewidth}
    \includegraphics[width=0.8\textwidth]{figures/logit_and_adv.png}
    \caption{Logit order and frequency of adversarial classification}
    \label{fig:logit_and_adv}
    \end{subfigure}
    \begin{subfigure}[b]{0.34\linewidth}
    \end{subfigure}
\caption{(\textbf{\subref{fig:logit_and_adv}}) Number of instances where the prediction of an adversarial sample corresponds to its $i$-th largest logit in CIFAR-10 %, sorted by the margin computed according to Def~\ref{def:cm} 
(excluding samples for which the prediction does not change). }
\label{fig:multiclassmargin}
\end{figure}
\fi
%
In Fig.~\ref{fig:multiclassmargin} we explore the relationship between the logits of a network evaluated at a clean sample and the predicted class of the adversarially perturbed variant. Methods which rely on the canonical notions of margin reasonably assume that samples at which a classifier is vulnerable have small margin according to Def.~\ref{def:cm}, i.e. the magnitude of the smallest difference between the logits of any class and the logit corresponding the true class is small. However, we demonstrate in  Fig.~\ref{fig:multiclassmargin}(b) that a significant number of predictions made by vulnerable classifiers on perturbed samples do not correspond to the classes with minimal margin. In other words, the class for which the margin is smallest does not always correspond to the adversarial class. Furthermore, this issue is exacerbated for robust networks as shown by the difference in count distribution between networks whose relative robustness varies. 

%As a consequence, a more reasonable assumption is that good weights are a function of a \emph{multi-class margin}\textemdash i.e. are both aware of the margin associated with each class as well as the true class associated with the sample.
%In this paper, we propose to learn a parametric model of the sample weights as a function of the multi-class margin of a sample.
%Weights are iteratively learned alongside the neural network parameters using a bilevel optimization scheme. 

%In this work, we propose to parameterize this function using a small \textit{auxiliary network}\textemdash i.e. the weights of the $i$-th sample are computed as $w_i = \omega_ \mu(\Delta_i)$, where $\mu$ are the parameters of the auxiliary network and trained via gradient descent. In general, we denote the function used to compute the weights $\omega(\cdot)$.

\subsection{Bi-level Optimization and Meta-learning}
Bilevel optimization, first introduced by \citet{braken1973bilevel} is an optimization framework involving nested optimization problems. 
A typical bilevel optimization problem takes on the form:
\begin{equation}
\min_{x \in \mathbb{R}^p} \Phi(x) := f(x, y^*(x))\:\: \textrm{s.t.}\:\: y^*\in \argmin_{y \in \mathbb{R}^p}g(x,y),
\end{equation}
where $f$ and $g$ are respectively denoted the \textit{upper-level} and \textit{lower-level} objectives. The goal of the framework is to minimize the primary objective $\Phi(x)$ with respect to $x$ where $y^*(x)$ is obtained by solving the lower-level minimization problem. 
The framework of bilevel optimization has seen adoption by the machine learning community\textemdash in particular in the context of hyperparameter tuning \citep{simon2018reweighting, ren18l2rw} and meta-learning \citep{finnmaml17,rajeswaran2019metalearning}. 
Our proposed algorithm has some similarity to meta-learning \citep{finnmaml17, rusu2018metalearning, rajeswaran2019metalearning, eshratifar2018gradagree}. Notably, the Model-Agnostic Meta-Learning (MAML) algorithm~\cite{finnmaml17} incorporates gradient information for the meta-learning setting. The application of meta-learning as an instance of bilevel optimization has been explored in the context of sample reweighting. For example, \citet{ren18l2rw, simon2018reweighting}, and \cite{han2018coteaching} proposed methods for learning with noisy labels by reweighting the gradients associated with the losses at individual samples based on balancing performance on a curated validation set and the corrupted training set.
\section{\algname{}: Learning Samples Weights for Adversarial Training}
\label{sec:method}
\begin{figure*}[t]
\centering
\includegraphics[width=\textwidth]{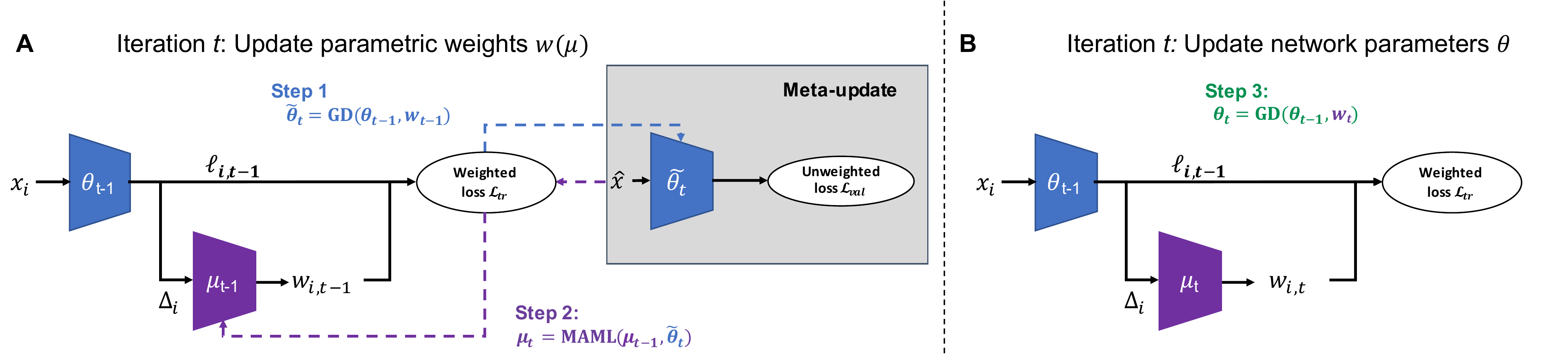}
\caption{\textbf{\algname{}} Framework. (\textbf{A}) 
Sample weighting. Step 1: intermediate parameters $\tilde{\theta}_t$ are computed by pseudo-update of $\theta_t$. Step 2: Validation loss gradients (calculated via back-propagation through the weighted training loss) are used to update the parameters of the auxiliary weighting network $\mu_t$. (\textbf{B}) Step 3: network parameters $\theta_t$ updated using new weights $w_t$.}
\label{fig:mml}
\end{figure*}
In this section, we propose \algname{}, a new learning framework for robust training. There are two main novelties in our new learning scheme compared to existing robust training methods. First, we consider a more reasonable assumption leveraging the concept of \emph{multi-class margin} in robust training, where good weights should be aware of both the margin associated with each class, as well as the true class associated with the sample. Second, as opposed to related work which defines an explicit formula (based on approximation or heuristics) for the weights dependent on the margin, we propose to \emph{learn} the weights as part of training the classification model. Specifically, we define the weights as a function of a multi-class margin, and parameterize this function using a small {auxiliary network}. We formulate this as a bi-level optimization problem and learn the weights iteratively with the classifier parameters. %In the following, we introduce the two key components in \algname{}: Multi-class margin reweighting in Sec 3.1 and Bilevel optimization formulation in Sec 3.2. 

% \lily{We illustrate xxx in Sec 3.1, xxx in Sec 3.2.}

% In this case, the Weights are iteratively learned alongside the neural network parameters using a bilevel optimization scheme. 

% To counter the issues arising in margin-based weighting, a more reasonable assumption is that good weights are a function of a \emph{multi-class margin}\textemdash i.e. good weights should be aware of both the margin associated with each class, as well as the true class associated with the sample.

\subsection{Multi-class Margin Reweighting}
We extend the logit-based definitions of margin applied in \citet{zeng2020adversarial, wang2021probabalisticmarginat} and define the multi-class margin of a classifier at a sample as follows.
\begin{definition}\label{def:mcm} The multi-class margin of a classifier $f$ on sample $(x_i
, y_i)$, denoted $\Delta : [0,1]^k \to [-1,1]^k$, is a $k$-dimensional vector whose $j$-th entry, $\Delta^{(j)}(f(x_i;\theta), y_i)$, is the difference between the classifier’s confidence in the correct label $y_i$ and the classifier’s confidence in label $j$, $\Delta^{(j)}(f(x_i;\theta), y_i) = p(f(x_i;\theta) = y_i) - p(f(x_i;\theta) = j)$.
\end{definition}
For brevity we denote $\Delta(f(x_i;\theta), y_i)$ as $\Delta_i$. Note that the multi-class margin exhibits two qualities:

\noindent 1. Correct/incorrect classification is implicit as negative values indicate an incorrect classification. 

\noindent 2. The true class of the sample is also implicit\textemdash i.e. the index with element zero (assuming the sample does not lie exactly on a decision boundary separating the true class from another).

In particular, we highlight the second quality. Prior work has demonstrated that the distribution of predictions made on adversarial samples is not necessarily uniform over all classes \citep{Abbasi2017RobustnessTA}. In other words, vulnerable samples and their associated adversarial perturbations may concentrate about certain classes more than others. 
We demonstrate in the results %This is demonstrated in Fig.~\ref{fig:cifarexamples}b where we demonstrate 
that networks exhibit non-uniform robustness per-class.

To learn the sample weights as a function of the multiclass margin, we construct an auxiliary neural network with a single hidden layer, whose parameters are denoted $\mu$ and whose inputs are the multi-class margins. % and the weights are the outputs of the netowrk, i.e.,
The weight of the $i$-th training sample is then computed as $w_i = \omega_ \mu(\Delta_i)$.
In general, we denote the function used to map from margin to weight $\omega(\cdot)$.
A question that arises is what loss function should be used to train this auxiliary network. We design a bilevel optimization approach leveraging the validation set to learn the auxiliary network parameters $\mu$.

\subsection{Bilevel Optimization}

%\subsection{Bilevel optimization for reweighting} %\lily{change subsection name: bi-level opt for reweighting?}
%\lily{Give $w$ and $W$ a name in the algorithm pseudo code?}
% 
We exploit a validation set to jointly learn a parametric weighting function $\omega_\mu$ on the training samples and a classifier which jointly minimize the associated weighted robust error.
Let $\hat{\mathcal{L}}_{\textrm{tr}}(\theta_t, w_t) = \sum_{i=1}^{m}w_{t,i}\hat{\ell}_i(\theta_t)$, where $w_{t,i} = \omega(\Delta_i;\mu_t)$ is the \textit{weighted} robust training loss with respect to parameters $\theta_t$ and $\mu_t$ at time $t$. Additionally, $w_{t,i} \geq 0$ and $\sum_{i=1}^{m_b} w_{t,y}=1$.  
Intuitively, the samples with high weights should improve robust generalization\textemdash this is quantified by the robust error evaluated on a held-out validation set. Let $\hat{\mathcal{L}}_{\textrm{val}}(\theta_t) = \frac{1}{n}\sum_{i=1}^{n}\hat{\ell}_i(\theta_t)$ be the \textit{unweighted} robust validation loss associated with $\theta_t$.
Following the meta-learning principle, we seek weights such that the minimizer of the weighted robust training loss maximizes robust accuracy on the unweighted validation set\textemdash i.e. solve the following bilevel optimization problem:
%
\iffalse
\begin{equation}
\begin{aligned}
&\arg \min_\theta \hat{\mathcal{L}}_{\textrm{tr}}(\theta, \omega(\Delta;\mu^*)) \\
&\:\textrm{s.t. } \mu^* \in \arg \min_\mu \hat{\mathcal{L}}_{\textrm{val}}(\theta)
\end{aligned}
\end{equation}
\fi
%
\begin{equation}
\arg \min_\theta \hat{\mathcal{L}}_{\textrm{tr}}(\theta, \omega(\Delta;\mu^*))\quad \textrm{s.t. } \mu^* \in \arg \min_\mu \hat{\mathcal{L}}_{\textrm{val}}(\theta)
\end{equation}
We provide a high-level overview of the procedure in Fig~\ref{fig:mml} and the reweighting algorithm in Alg.~\ref{alg:training_proc}.
%The workflow of the reweighting procedure is illustrated in Fig. \ref{fig:mml}(A)   %The detailed description of the procedure is presented in Alg.~\ref{alg:training_proc}.

\begin{algorithm}[t]
\caption{\textbf{\algname{}} training procedure}
\begin{flushleft}
\textbf{Input:} Training data $\mathcal{D}$, validation-data set $\mathcal{\hat{D}}$, max iterations $T$, learning rates $\alpha, \beta$\\
\textbf{Output:} Classifier parameters $\theta$
\end{flushleft}
\begin{algorithmic}[1]
\State $t \gets 0$
\State Initialize $\theta_0$, $\mu_0$, $w_{0} = \omega_{\mu_{0}}(\Delta)$
\For{$t\leq T$}    
    \State $(X,y) \sim \mathcal{D}$, $(\hat{X},\hat{y}) \sim \mathcal{\hat{D}}$
    
\State $\tilde{\theta}_t \gets \theta_t - \beta\cdot\nabla_\theta\hat{\mathcal{L}}_\textrm{tr}|_{\edit{\theta_t}, w_t}$
    \State $\mu_{t+1} = \mu_t - \alpha\nabla_\mu\hat{\mathcal{L}}_{\textrm{val}}|_{\edit{\tilde{\theta}_t}, w_t}$ 
    \Comment{compute $\nabla_\mu\hat{\mathcal{L}}_{\textrm{val}}|_{\tilde{\theta}_t,w_t}$ via backpropagation according to Eq.~(\ref{eq:wupdate}})
    %\Comment{compute $\nabla_\mu\hat{\mathcal{L}}_{\textrm{val}}|_{\theta_t, w_t}$ according to Alg. \ref{alg:reweighting_proc}}

    \State compute $w_{t+1} = \omega_{\mu_{t+1}}(\Delta)$
    \Comment{compute $\Delta$ with respect to $\theta_t$ according to Def~\ref{def:mcm}}
    \State $\theta_{t+1} = \theta_{t} - \beta\nabla_{\theta}\hat{\mathcal{L}}_{\textrm{tr}}|_{\theta_t, w_{t+1}}$
\EndFor
\State \Return $\theta_T$
\end{algorithmic}
\label{alg:training_proc}
\end{algorithm}
Our approach is composed of three steps. Steps 1 and 2 rely on the MAML-trick \citep{finnmaml17}, which substitutes one-step updates $\mu_t$ for $\mu^*$ and iteratively solves the upper-level problem. In this context, $\mu_t$ is updated according to the gradient of the unweighted robust validation loss with respect to the sample weights. %(Step 2 in Fig.~\ref{fig:mml}A, line 5 in Alg.~\ref{alg:training_proc}). 
%We note that this method necessitates computation of a \textit{pseudo-update} for $\theta_{t-1}$ on which to compute the validation loss in order to compute the gradient updates for $\mu$:
We note that this method necessitates computation of a \textit{pseudo-update} in order to compute this gradient:

\textbf{Step 1} Pseudo update of classifier parameters $\tilde{\theta}_t$ (Step 1 in Fig.~\ref{fig:mml}, line 5 in Alg.~\ref{alg:training_proc})
\begin{equation}
\tilde{\theta}_t = \theta_{t} - \beta\nabla_\theta\hat{\mathcal{L}}_{\textrm{tr}}(\theta_{t}, w_{t-1})
\end{equation}
The pseudo parameters $\tilde{\theta}_t$ are then used as a surrogate for $\theta_t$ in optimizing $\mu$: %Eq.~\ref{eq:wupdate}.

\textbf{Step 2} Update parameters $\mu_t$ of the auxiliary network (Step 2 in Fig.~\ref{fig:mml}, line 6 in Alg.~\ref{alg:training_proc})
\begin{equation}
\begin{split}
\mu_{t} & = %\mu_{t-1} - \frac{\alpha\beta}{mn}\sum_{j=1}^{m}\left(\sum_{i=1}^{n}\left(\frac{\partial \hat{\ell}^{\textrm{val}}_i(\tilde{\theta})}{\partial \tilde{\theta}}\bigg\lvert_{\tilde{\theta}_t}\right)^\top\frac{\partial \hat{\ell}^{\textrm{tr}}_j(\theta)}{\partial \theta}\bigg\lvert_{\theta_{t-1}}\right)\frac{\partial w}{\partial \mu}\bigg\lvert_{\mu_{t}},
\mu_{t-1} - g,
\\
g & = \frac{\alpha\beta}{mn}\sum_{j=1}^{m}\left(\sum_{i=1}^{n}\left(\frac{\partial \hat{\ell}^{\textrm{val}}_i(\tilde{\theta})}{\partial \tilde{\theta}}\bigg\lvert_{\tilde{\theta}_t}\right)^\top\frac{\partial \hat{\ell}^{\textrm{tr}}_j(\theta)}{\partial \theta}\bigg\lvert_{\theta_{t-1}}\right)\frac{\partial w}{\partial \mu}\bigg\lvert_{\mu_{t}},
\label{eq:wupdate}
\end{split}
\end{equation}
where $\alpha$ and $\beta$ are the step size used in the pseudo and auxiliary network updates, respectively. 

\textbf{Step 3} Update parameters of classifier network (Step 3 in Fig.~\ref{fig:mml}, line 8 in Alg.~\ref{alg:training_proc})
\begin{equation}
\theta_{t+1} = \theta_{t} - \beta\nabla_\theta\hat{\mathcal{L}}_{\textrm{tr}}(\theta_{t}, w_{t})
\end{equation}
 One interpretation of this procedure is that we take a pseudo-step using $\theta_{t-1}$ and $\mu_{t-1}$ (Step 1), calculate the best update to auxiliary network parameters $\mu_t$ in \textit{hindsight} that improve generalization, by minimizing the validation loss with $\tilde{\theta}_t$, (Step 2), and then derive the ``true'' update for $\theta_{t-1}$ by minimizing the weighted training loss using the new weights $\mu_t$ (Step 3). The detailed derivation of the gradient update is provided in the appendix.
Note that the term
$\frac{1}{n}\sum_{i=1}^{n}\left(\frac{\partial \hat{\ell}^{\textrm{val}}_i(\tilde{\theta})}{\partial \theta}\bigg\lvert_{\tilde{\theta}_t}\right)^\top\frac{\partial \hat{\ell}^{\textrm{tr}}_j(\theta)}{\partial \theta}\bigg\lvert_{\theta_{t-1}}$ in Eq.~(\ref{eq:wupdate})
represents the correlation between the gradient of the $j$-th training sample computed on the training loss and the average gradient of the validation data calculated on the robust validation loss. As a consequence, if the gradient of the loss with respect to the network parameters at time $t$ for training sample $j$ is aligned with the average gradient of the meta-loss, it will be considered a beneficial sample for generalization and its weight will be increased. Conversely, the weight of the sample is suppressed if the gradient is anticorrelated with the average validation set-gradient. 

\iffalse
Under our framework, it is necessary to perform backpropagation three times: once to compute the meta-parameters, once to compute the meta-gradient, and once to perform the final update of the network parameters. In this instance, using iterative methods such as PGD to estimate the robust loss is prohibitively expensive. Recent work \cite{Wong2020Fast} has demonstrated that during training, adversarial perturbations crafted by taking a single step in the direction of the positive gradient is a viable substitute when the perturbations are initialized with uniformly random noise and early stopping is employed. Given an upper bound on the magnitude of the perturbation $\epsilon$, we construct an adversarial perturbation $\delta$ according to the following Fast-FGSM procedure:
%
\begin{align*}
\delta &= \textrm{Uniform}(-\epsilon,\epsilon) \\
\delta &= \delta + \alpha \cdot \textrm{sign}(\nabla_\delta f_\theta(x + \delta),y) \\
\delta &= \max(\min(\delta, \epsilon),-\epsilon)
\end{align*}
We apply this method to compute the empirical robust losses in our algorithm. 
\fi

%

%

%
\section{Experiments}
\label{sec:experiments}
In this section, we evaluate the efficacy of our framework on a variety of datasets, and demonstrate that our technique improves robustness while preserving clean accuracy. 
We introduce three variants based on our reweighting technique:

\noindent 1) Non-parametric reweighting: we learn weights using the \textit{weighted} adversarial cross-entropy loss where the weight $w_{j,t}$ for sample $j$ at iteration $t$ is proportional to the correlation between the training loss gradient and the average validation loss gradient: $\frac{1}{n}\sum_{i=1}^{n}\left(\frac{\partial \hat{\ell}^{\textrm{val}}_i(\tilde{\theta})}{\partial \theta}\bigg\lvert_{\tilde{\theta}_t}\right)^\top\frac{\partial \hat{\ell}^{\textrm{tr}}_j(\theta)}{\partial \theta}\bigg\lvert_{\theta_{t-1}}$.

\noindent 2)  \algname{} (Parametric reweighting, Sec.~\ref{sec:method}) trained using the \textit{weighted} adversarial cross-entropy loss. 

\noindent 3) \algname{}-TRADES: Parametric reweighting trained with the TRADES loss~\citep{trades}: %we solve
\begin{equation*}
    \min_{\theta}\sum_i\ell(f(x_i\mu_t;\theta), y_i) + 1/\lambda (w_i\cdot \textrm{KL}(f(x_i;\theta), f(x_i + \delta; \theta))),
\end{equation*}
where $\ell(\cdot)$ corresponds to the standard cross-entropy loss, $\textrm{KL}$ corresponds to the KL-divergence, $\delta$ corresponds to an adversarially perturbation, and $w_i = \omega(\Delta_i;\mu)$: the parametric map applied to the multi-class margin of $f_\theta$ at $x_i$. 
For all experiments, we set $1/\lambda = 6$, and define $\omega$ to be a single hidden-layer fully connected ReLU network with $128$ hidden units and a sigmoid activation. Furthermore, to enforce aforementioned constraints, we normalize the weights per-batch for all methods\textemdash i.e. $w_i = w_i / \sum_jw_j$.

\subsection{Performance evaluation}
\begin{table*}[ht]
\caption{CIFAR-10 comparison for AT, GAIRAT, WMMR, MAIL, and \algname{} variants with standard adversarial training (\algname{}) and TRADES loss (\algname{}-TRADES). 
We report clean test accuracy, PGD, and AutoAttack (AA) robust accuracy. We perform AA on $1000$ samples. The \textbf{\underline{best}} result is underlined \& bolded and \textbf{second best} is bolded. We emphasize performance on the last column.}
\label{tab:resultscifar}
\centering
\resizebox{0.95\textwidth}{!}{%
\begin{tabular}{|l| l l l||l l l||l l l||l l l|} 
\hline

\multirow{2}{*}{}    & \multicolumn{6}{c||}{ Small-CNN } & \multicolumn{6}{c|}{ WRN-10-32 }     \\ 

\cline{2-13}

\multirow{2}{*}{} & \multicolumn{3}{l||}{perturbation: $\ell_\infty$ } & \multicolumn{3}{l||}{perturbation: $\ell_\infty$ }      & \multicolumn{3}{l||}{perturbation:~$\ell_\infty$ }   & \multicolumn{3}{l|}{perturbation:~$\ell_\infty$}  \\ 

%\cline{2-13}

       & Clean    & PGD    & AA         & Clean    & PGD   & AA              & Clean    & PGD    & AA            & Clean    & PGD    & AA               \\ 
\hline
\hline

\textit{CIFAR-10}       & \multicolumn{3}{c||}{ $\epsilon = 0.0078$}  & \multicolumn{3}{c||}{ $\epsilon = 0.031$ } & \multicolumn{3}{c||}{ $\epsilon = 0.0078$ }    & \multicolumn{3}{c|}{ $\epsilon = 0.031$ }        \\ 

\hline
GAIRAT   &  {79.0}  & 54.7  &   48.1      & 79.0   & \textbf{55.6}   &     40.7       &  \textbf{86.4}  &   73.6   &  63.1   & 84.7  & 56.8  &   43.4            \\ 
%\hline
WMMR   &  78.7 &  58.9  &    51.2      &  \textbf{81.7} & 49.1   &     39.1       &  85.9   & 70.9  &  67.4  &       80.6 & 49.5   &   40.6           \\ 
MAIL  &  76.8  & \textbf{64.3}   &    \textbf{59.2}      & \textbf{\underline{81.9}}   & 53.3     &   40.6        & 84.3   & 74.1  &\textbf{73.7}  &   83.2   &  53.7      & {52.0}              \\  
\hline
AT    & 78.7  &  58.7   &    56.6  & 79.6     & 45.6    &    42.9     & 85.9    & 71.3   &    69.5    & 85.9  & 52.0  &     48.0        \\  %\hline
TRADES ($1/\lambda = 6$) & \textbf{79.2}   & 58.9   &   56.8      &  78.9   &  54.8   &    \textbf{51.7}      & 84.6   & 73.9   & 73.1   &    83.1     &  53.9 & \textbf{52.1}            \\  
Non-parametric weighting    &  \textbf{\underline{79.7}}  & 60.0   &  47.3        & 81.3   &  52.2   &    40.6     &  \textbf{86.4}  & 73.7  & 62.3  &     {86.6}    & 52.8 &  42.9           \\  
\algname{}  (ours)  &  \textbf{\underline{79.7}}  & 63.6   &      56.7    &  80.4  & 55.4   &  45.3      &  \textbf{\underline{87.1}}   & \textbf{74.2}  & 71.3 & \textbf{\underline{87.4}}   & \textbf{57.2} & 51.4           \\  
\algname{}-TRADES (ours) &  {79.1}  & \textbf{\underline{64.8}}   &    \textbf{\underline{61.5}}      &  80.2  & \textbf{\underline{56.2}}     &     \textbf{\underline{52.6}}     &    {86.2} & \textbf{\underline{74.8}}   &  \textbf{\underline{74.2}}  &    \textbf{87.1}     & \textbf{\underline{57.4}}  &  \textbf{\underline{53.6}}           \\  
\hline
\end{tabular}}
\end{table*}
We evaluate the performance of our approach compared to plain training, adversarial training (AT)~\citep{madry2018towards}, GAIRAT~\citep{zhang2021geometryaware}, WMMR~\citep{zeng2020adversarial}, and MAIL~\citep{wang2021probabalisticmarginat}. All experiments are run on a single RTX 2080 Ti. When applying our approach and variants, two validation sets of size $1000$ are extracted from the training set: one is used to learn the auxiliary network parameters, and the second is used for early stopping. This results in a smaller training set for \algname{}, while the training sets of competing methods are unaltered.
%
%
%We note two potential drawbacks of our algorithm in this context: (1) our dataset requires a re-partitioning of of the training set, which reduces the number of samples seen by the classifier. (2) the MNIST and Fashion-MNIST datasets contain a non-trivial number of misclassified samples which can influence performance~\citep{mislabeled}. In the case of algorithms which perform weighted training, the occurrence of highly weighted outliers or mislabeled samples is a danger. Addressing the issues of sample complexity and label noise is future work.
In Table~\ref{tab:resultscifar}, we evaluate our method using the two architectures used in~\citet{zhang2021geometryaware} on CIFAR-10~\citep{KrizhevskyCIFAR10}: a 6-layer convolutional network (Small-CNN) and a Wide-Resnet-32-10 (WRN-32-10)~\citep{zagoruykowrn16}, with details provided in the Appendix. We run each method for 100 epochs with training and validation batch sizes set to 128 using SGD + momentum. A standard learning rate schedule is implemented with the initial
learning rate of 0.1 divided by 10 at Epoch 30 and 60, respectively. 
We consider robustness with respect to $\ell_\infty$ distance. We report three criteria: clean test accuracy (clean), robust test accuracy (PGD), and AutoAttack (AA). Robust test accuracy is computed using Projected Gradient Descent (PGD)~\citep{madry2018towards} with 20 iterations.
\begin{wraptable}[11]{l}{75mm}
\caption{CIFAR-100 comparisons between baselines + TRADES and BiLAW + TRADES. *=reported result.}
\label{tab:resultscifar100}
\centering
\begin{tabular}{|l| l l l|} 
\hline

%\multirow{2}{*}{}    & \multicolumn{3}{c|}{ WRN-10-32 }     \\ 

\cline{2-4}

WRN-32-10       & Clean    & PGD    & AA     \\
\hline
\hline

\textit{CIFAR-100}       & \multicolumn{3}{c|}{$\ell_\infty$  $\epsilon = 0.031$}         \\ 

\hline
%AT  &  57.9  & 28.9  & 24.7 \\
TRADES ($1/\lambda=1$)    &  \textbf{62.4}  & 25.3  & 22.2 \\
TRADES ($1/\lambda=6$)    &  56.5  & 30.9  & \textbf{26.9} \\
BiLAW-TRADES (ours)   &  \textbf{\underline{62.8}}  & \textbf{31.4}  & \textbf{\underline{27.2}}  \\
GAIR-TRADES  & 61.4  &  \textbf{\underline{32.7}}  & 23.4  \\
MAIL-TRADES*   & 60.1  & 30.3 & 24.8 \\ \hline
\end{tabular}

\end{wraptable}
%\looseness=-1

\textbf{\algname{}} strictly outperforms AT with respect to both clean and robust accuracy and generally outperforms GAIRAT and WMMR with respect to clean and robust accuracy on CIFAR-10 (up to $10\%$). In particular, \algname{} consistently achieves superior clean test accuracy in all testcases, except for the $\ell_\infty$ small-CNN ($\epsilon = 0.031$). On the WRN $\ell_\infty$ case, we maintain and outperform relevant methods with respect to both PGD-based and AA-based robust accuracy while achieving superior clean test accuracy. 
We demonstrate that when used in conjunction with TRADES, \algname{} preserves and improves robustness to AA attacks by $1.5\%$ in contrast to TRADES, while significantly enhancing clean test accuracy by up to $3\%$ and PGD attacks by up to $5\%$. 
We also note that parametric reweighting as opposed to non-parametric reweighting significantly improves robust accuracy.  
On CIFAR-100 (Table~\ref{tab:resultscifar100}) \algname{}-TRADES out-performs all other methods with respect to clean and AA-based robust accuracy. Our results demonstrate the effectiveness of using a held-out validation set to learn the sample weights compared to heuristic reweighting schemes.

In Table~\ref{tab:results} in Appendix~\ref{app:mnist}, we evaluate \algname{} using two smaller networks on MNIST~\citep{lecun-mnisthandwrittendigit-2010} and Fashion-MNIST~\citep{Xiao17}. In all testcases, \algname{} matches the performance of GAIRAT, out-performs the other reweighting methods for clean, PGD, and AA accuracy. %However, we note the overall distribution of both clean and robust accuracy is tight.
%
%In the Appendix we conduct three ablative  experiments to analyze the effect of (1) the capacity of the auxiliary weighting network and its ability to generalize, (2) the fixed TRADES coefficient and (3) the input to the weighting network, on the performance of BiLAW.
%
In the Appendix we conduct two ablative experiments to analyze the effect of (1) the TRADES coefficient and (2) the input encoding to the weighting network. We also show that F-FGSM~\cite{Wong2020Fast} may be used to improve the efficiency of \algname{}.\looseness=-1

\subsection{Robustness to weight-aware adversaries}

We investigate the question: \textit{are classifiers trained with reweighting robust to adversaries that have partial or complete knowledge of the reweighting mechanism?} %We provide affirmative evidence.
We discuss two instantiations of a weight-aware adversary: (1.) an adversary which treats the weights as constants and (2.) treats the weights as a function of the classifier and labels. If the $w_i$ are considered constants, the optimal adversarial perturbation will be the same regardless of knowledge of $\omega$. Consider an untargeted attack:\looseness=-1
\begin{equation}
\max_{\delta} \ell(f(x_i + \delta; \theta),y_i) 
\label{eq:untargeted}
\end{equation}
where $x_i$ is the original image, $y_i$ is the associated label, $\delta$ is a perturbation subject to the constraints $||\delta||_p \leq \epsilon$, $x_i +\delta \in [0,1]^n$, and $\theta$ are the classifier parameters. A weight-aware adversary solves%In a weight-aware attack, the attacker solves
\begin{equation}
\max_\delta \omega(\Delta(f(x_i + \delta; \theta), y_i);\mu)\cdot \ell(f(x_i + \delta; \theta),y_i) 
\label{eq:adaptive}
\end{equation}
The solutions of the two problems are the same as long as the weight is positive (which is guaranteed via a normalization layer).
On the other hand, if each $w_i$ is treated as a function of the classifier, an attacker could indeed perform gradient ascent on the loss suffered by the classifier at an input. In particular, the gradient of the perturbation would be decomposed into the sum of two parts:

    \noindent 1. The typical adversarial direction scaled by the predicted weight (a function of the margin): $\omega(\Delta(f(x_i + \delta; \theta), y_i);\mu)\frac{\partial}{\partial \delta}\ell(f(x_i + \delta; \theta),y_i)$
    
    \noindent 2. The gradient of the weighting network with respect to the perturbation:\\ $\ell(f(x_i + \delta; \theta),y_i)\frac{\partial}{\partial \delta}\omega(\Delta(f(x_i + \delta; \theta), y_i);\mu)$

%This implies that the solution would be different compared with the solution to Eq.~(\ref{eq:untargeted}) $\max_{\delta} \ell(f(x_i + \delta; \theta),y_i)$ and Eq.~(\ref{eq:adaptive}) $\max_\delta w_i\ell(f(x_i + \delta; \theta),y_i)$. However, we claim that \textit{the weight-aware attack will only give an \textbf{equal} or \textbf{worse} solution to Eq.~(\ref{eq:untargeted})} due to the fact that Eq.~(\ref{eq:untargeted}) is the \textit{true} formulation of the adversarial perturbation, while Eq.~(\ref{eq:adaptive}) \textit{is not}\input{tables/ablation2}
This implies that the solution would be different compared with the solution to Eq.~(\ref{eq:untargeted}) and Eq.~(\ref{eq:adaptive}). 
\begin{wraptable}[10]{l}{80mm}
\vspace{-0.25cm}
\caption{Clean and robust accuracy of weight-adaptive adversaries. Lower implies a stronger attack.}%in terms of clean and robust (PGD) accuracy.}
\label{tab:ablation2}
\centering
%\resizebox{0.4\textwidth}{!}{%
\begin{tabular}{|l| c c|} 
\hline
%\multirow{2}{*}{}    & \multicolumn{2}{c|}{ SMALL-CNN }     \\ 

\cline{2-3}

%\multirow{2}{*}{} & \multicolumn{2}{l|}{$\ell_\infty$ $\epsilon = 0.031$}  \\ 

%\cline{2-13}

 SMALL-CNN      & Clean    & PGD       \\
\hline
\hline

\textit{Adversary}       & \multicolumn{2}{c|}{$\ell_\infty$ $\epsilon = 0.031$}         \\ 

\hline
no knowledge (Eq.~\ref{eq:untargeted}) &  80.2  & 56.2  \\
partial knowledge (Eq.~\ref{eq:adaptive})  & --  & 56.8   \\
full knowledge (Eq.~\ref{eq:adaptive})  & -- & 	57.3  \\
\hline
\end{tabular}%}
\end{wraptable}However, we claim that \textit{the weight-aware attack will only give an \textbf{equal} or \textbf{worse} solution to Eq.~(\ref{eq:untargeted})} due to the fact that Eq.~(\ref{eq:untargeted}) is the \textit{true} formulation of the adversarial perturbation, while Eq.~(\ref{eq:adaptive}) \textit{is not}.
Thus, if an attacker solves Eq.~(\ref{eq:adaptive}) to perform a weight-aware attack (i.e. with knowledge of the sample weights), it's actually \textit{harmful} to the attack performance.\looseness=-1
To support our argument, we perform a weight-aware attack on the Small-CNN classifier. In the true white-box setting, an attacker may have access to the true weights of the weighting network. However, it is more likely that an attacker may only have knowledge of the usage of the BiLAW framework during training. In this case, an attacker might be able to train a weighting network independently or utilize a pre-trained weighting network. We evaluate an attacker that has full knowledge of the weighting network and an attacker which only has access to a pre-trained weighting network (partial knowledge) in Table~\ref{tab:ablation2} and find that the weight-aware attack result is slightly worse than the standard attack result. These results imply that even oracle knowledge of the weighting network does \textbf{NOT} help an attacker, thus justifying our statement. \looseness=-1

\vspace{-0.2cm}
\subsection{Training sample weights}
\begin{figure}[!t]
    \centering
\includegraphics[width=0.9\linewidth]{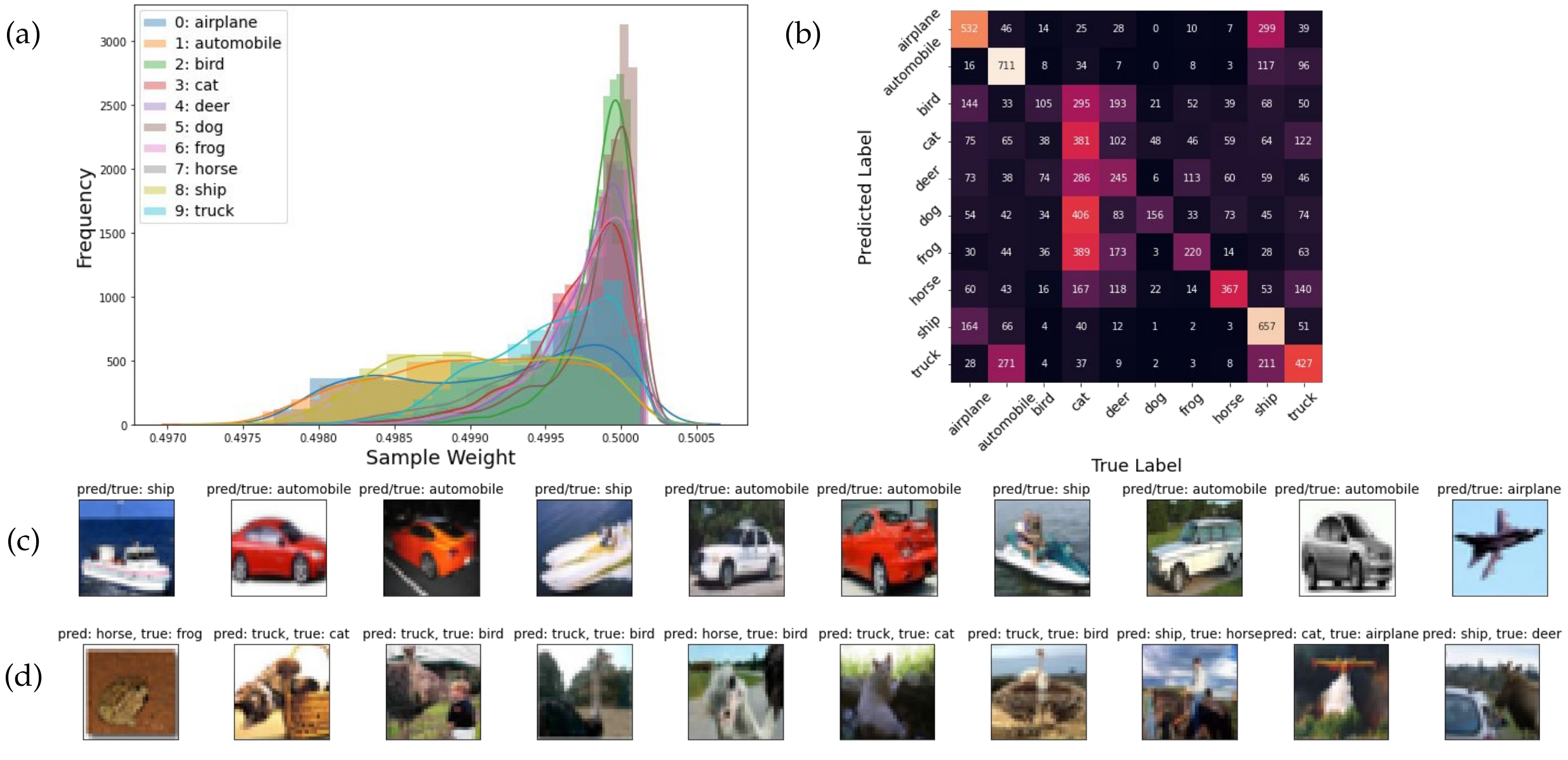}
%\vspace{-0.3cm}
\caption{(\textbf{a}) Weight distribution of CIFAR-10 samples. (\textbf{b}) Adversarial confusion matrix of a robust network on CIFAR-10.
(\textbf{c}) ``Easy'' CIFAR-10 samples with low weight are correctly classified.
(\textbf{d}) ``Hard'' CIFAR-10 samples with high weight are typically incorrectly classified.}
\label{fig:cifarexamples}
\vspace{-0.75cm}
\end{figure}
We investigate the correspondence between weights and samples, and ask the question: \textit{what are the properties of training examples with high/low weights?} 
Fig.~\ref{fig:cifarexamples} provides evidence that supports our claim that samples for which the auxiliary network predicts high weights correspond to vulnerable, or difficult samples close to the decision boundary. 
In Fig.~\ref{fig:cifarexamples}(a)-(b), we plot the distribution of weights for each class, as well the associated confusion matrix of predictions made by a robust classifier (trained with \algname{}) on adversarial samples. 
We note that the distribution of weights matches the distribution of misclassified adversarial examples. For example, in Fig.~\ref{fig:cifarexamples}(a), samples of the `ship' and `automobile' classes are assigned a higher number of smaller weights and they are typically classified correctly as in Fig.~\ref{fig:cifarexamples}b. In contrast, birds, cats, and other animals have a higher number of samples assigned large weight and are more frequently misclassified. 

%In Fig.~\ref{fig:cifarexamples}(c)-(d) we provide several examples of test samples with low and high weights, respectively.
In Fig.~\ref{fig:cifarexamples}(c), we provide several examples of test samples that are assigned low weight. These images typically involve a centered object and plain background. In Fig.~\ref{fig:cifarexamples}(d), we provide a set of test samples assigned high weights. Many of these images are challenging for humans to identify, even when uncorrupted by adversarial noise. For example, the second and fifth image are pictures of cats and birds with unusual pose. The seventh, eighth, and ninth image are nearly impossible to identify due to complex backgrounds or obscured objects. Additionally, the second, third, eighth, and tenth images consist of multiple objects that could confuse the network or facilitate more effective perturbations.
%Thus, high weights can be used to identify adversarially vulnerable samples, and automatically differentiate easy and challenging samples in existing datasets.

%
\section{Conclusion}
\label{sec:conclusion}
We have introduced \textbf{\algname{}}, a new robust training method to train a robust classifier via \textit{learned} sample weights. We demonstrate that our method learns robust networks that out-performs competing methods, including recently proposed margin-aware adversarial training techniques. Notably, \textbf{\algname{}} does not rely on complicated heuristics to assign weights, and we demonstrate the learned weights are interpretable. Future work involves improving scalability and investigating whether the auxiliary network might be used to \textit{detect} adversarial corruptions. 
%Furthermore, the scalability of our method can be improved by substituting one-step adversarial perturbations for PGD during training.

\bibliography{main}
\newpage
\section{Appendix}
First, we review the derivation of the meta gradient in Sec.~6.1. In Sec.~6.2, we provide the architecture and training parameters for our experiments. In Sec.~6.3\textemdash 6.6, we provide additional experiments to highlight the effect of 1. the capacity and input of the weighting network on the clean and robust test accuracy, 2. more efficient variants of PGD on performance and runtime, 3. a comparison with TRADES with different weight parameter, 4. the sample weights\textemdash examples of training samples assigned large and small weights and the correlation of weights computed using \algname{} with weights produced by related approaches. 
\subsection{Derivation of Meta Gradient}
In this section we derive the update rule for the parameters of the auxiliary network in Eq.~\ref{eq:wupdate}:
\begin{equation*}
\mu_{t} = \mu_{t-1} - \frac{\alpha\beta}{mn}\sum_{j=1}^{m}\left(\sum_{i=1}^{n}\left(\frac{\partial \hat{\ell}^{\textrm{val}}_i(\tilde{\theta})}{\partial \tilde{\theta}}\bigg\lvert_{\tilde{\theta}_t}\right)^\top\frac{\partial \hat{\ell}^{\textrm{tr}}_j(\theta)}{\partial \theta}\bigg\lvert_{\theta_{t-1}}\right)\frac{\partial w}{\partial \mu}\bigg\lvert_{\mu_{t}},
\label{eq:wupdate_apdx1}
\end{equation*}
Let 
$$\hat{\mathcal{L}}_{\textrm{tr}}(\theta_t, w) = \frac{1}{m}\sum_{j=1}^{m}w_j\hat{\ell}_j(\theta_t)$$ be the robust training loss with respect to parameters $\theta$ at time $t$ and example weight $w_j$ for the $j$-th training example. Let $\hat{\mathcal{L}}_{\textrm{val}}(\theta_t) = \frac{1}{n}\sum_{i=1}^{n}\hat{\ell}_i(\theta_t)$ be the associated \textit{unweighted} validation loss. Following the meta-learning framework, we to minimize this loss via gradient descent.
\begin{align*}
\frac{\partial \hat{\mathcal{L}}_\textrm{val}(\tilde{\theta})}{\partial \mu} &= \frac{1}{n}\sum_{i}^{n}\frac{\partial \hat{\ell}^{\textrm{val}}_i(\tilde{\theta})}{\partial \mu} \\
&= \frac{1}{n}\sum_{i}^{n}\frac{\partial \hat{\ell}^{\textrm{val}}_i(\tilde{\theta})}{\partial \tilde{\theta}} \frac{\partial \tilde{\theta}}{\partial w}\frac{\partial w}{\partial \mu}
\end{align*}
To compute $\frac{\partial \tilde{\theta}}{\partial w}$, we can apply the MAML technique and differentiate through the pseudo update (recall, $\tilde{\theta}_t = GD_{\textrm{tr}}(\theta_{t-1}, w_{t-1}) := \theta_{t-1} - \alpha\nabla_{\theta}\mathcal{L}_{\textrm{tr}, t-1}(\theta_{t-1},w)$). For example, a single gradient descent step:
\begin{align*}
\frac{\partial \tilde{\theta}}{\partial w} &= \frac{\partial}{\partial w} (\theta_{t-1} - \alpha\nabla_{\theta}\mathcal{L}_{\textrm{tr}, t-1}(\theta_{t-1},w)) \\
&=\left(\frac{\alpha}{m}\sum_{i=1}^{m}\nabla_{\theta}\hat{\ell}^{\textrm{tr}}_{t-1}(\theta_{t-1})\right)
\end{align*}
So the complete update is:
\begin{equation*}
\mu_{t} = \mu_{t-1} - \frac{\alpha\beta}{mn}\sum_{j=1}^{m}\left(\sum_{i=1}^{n}\frac{\partial \hat{\ell}^{\textrm{val}}_i(\tilde{\theta})}{\partial \tilde{\theta}}\bigg\lvert_{\tilde{\theta}_t}^\top\frac{\partial \hat{\ell}^{\textrm{tr}}_j(\theta)}{\partial \theta}\bigg\lvert_{\theta_{t-1}}\frac{\partial w}{\partial \mu}\bigg\lvert_{\mu_{t}}\right)
\label{eq:wupdate_apdx2}
\end{equation*}

\subsection{Experiments}

\subsubsection{Architectures}
\label{app:architecture}
\begin{table}[h]
\caption{Architectures for main experiments for number of classes $nc$.}
\label{tab:arch}
\begin{center}
\begin{tabular}{c|c|c|}
FC1          & tiny-CNN   & small-CNN                                \\
\hline 
                                     &                                       \\
FC($1024$)           & Conv($16$, $4 \times 4$, $2$)    &    small-CNN-BLOCK($64$)                   \\
ReLU                 & ReLU                             &    small-CNN-BLOCK($128$)                     \\
FC($nc$)             & Conv($32$, $4 \times 4$, $2$)    &    small-CNN-BLOCK($196$) \\
                     & ReLU                             &    FC($256$) \\
                     & FC($100$)                        &    ReLU  \\
                     & ReLU                             &    FC($nc$)  \\
                     & FC($nc$)                         &  \\
\end{tabular}
\end{center}
\vspace{-5mm}
\end{table}

\begin{table}[h]
\caption{Architectures for main experiments for number of classes $nc$.}
\label{tab:arch}
\begin{center}
\begin{tabular}{|c|}
small-CNN-BLOCK($c$)                               \\
\hline 
Conv($c$, $3 \times 3$, $1$)   \\
BatchNorm \\
ReLU \\
Conv($c$, $3 \times 3$, $1$) \\
BatchNorm \\
ReLU \\
MaxPool($2\times 2$)  \\
\end{tabular}
\end{center}
\vspace{-5mm}
\end{table}

We abbreviate one hidden layer fully connected network with 1024 hidden units with FC1. The tiny-CNN convolutional architecture that we use is identical to that of \cite{KolterWongPolytope17, Croce2020Provable} —consisting of two convolutional layers with $16$ and $32$ filters of size $4 \times 4$ and stride $2$, followed by a fully connected layer with $100$ hidden units. For all experiments we use training and validation batch sizes of $128$ and we train all models for $100$ epochs. Moreover, we use SGD with a piecewise constant learning rate schedule with initial learning rate of $0.1$. The learning rate is divided by $10$ at epochs $30$ and $60$ respectively. On all datasets (MNIST, F-MNIST, CIFAR-10, and CIFAR-100) we restrict the input to be in the range $[0, 1]$. On the CIFAR-10 dataset, following \cite{zhang2021geometryaware}, we apply random crops and random mirroring of the images as data augmentation during training. We perform adversarial training using the PGD attack of \cite{madry2018towards}. During training, we perform 10 iterations of the PGD attack for all datasets. During evaluation, we use 20 iterations for all datasets. Following \cite{zhang2021geometryaware}, the step size is the perturbation radius divided by $4$.

\subsection{Capacity and generalization}
We explore how the capacity of the auxiliary reweighting technique influences the performance of our method. We also demonstrate the advantage of the multi-class margin over alternative inputs mapping to the sample weights\textemdash e.g. using the class-unaware margin (Def.~1), the adversarial loss $\Delta_\textrm{adv}$, and the difference between the adversarial loss and the clean loss at a sample $\Delta_\textrm{diff}$.

\begin{table}[h!]
\caption{Capacity of the auxiliary weight prediction network}
\label{tab:ablat_capacity}
\centering
\begin{tabular}{|l|c|c|c|}
\hline
\multirow{2}{*}{Capacity of $\omega$} &  \multicolumn{3}{c|}{CIFAR10} \\ \cline{2-4} 
                  & Clean           & PGD  & PGD - Clean             \\ \hline
$64$             &    83.6      &    57.4   & 26.2    \\ 
$64-64$             &    85.8     &   57.6  & 28.2  \\ 
$128$               &    87.1      &    57.4   & 27.7    \\ 
$256$             &    85.7     &     57.7   & 29.4     \\ 
 \hline \hline 
pretrained (128)              &     86.4     &   56.2    &   30.2 \\
 \hline
\end{tabular}
\end{table}
In Table~\ref{tab:ablat_capacity}, we evaluate the influence of the auxiliary network architecture and capacity, i.e. the choice of $\omega$. We observe that the architecture of the network influences the clean-robust tradeoff, with smaller networks (64 hidden units) reducing the gap between clean and robust performance, and larger networks (256 hidden units) increasing the gap. 

Furthermore, we demonstrate the feasibility of leveraging a pretrained reweighting network. We first train a robust classifier (Small-CNN) with a reweighting network using BiLAW. We then train a new  WRN classifier to minimize the weighted robust TRADES loss, where the sample weights are determined by the fixed, pretrained weighting network. Note that in this setting, the weighting function is no longer updated and the cost of training is equivalent to standard backpropagation (with a forward pass through the pretrained weighting network to compute the sample weights). As expected, we observe a minor degradation in clean and robust accuracy. However, the performance matches or exceeds that of the heuristic weighting functions (WMMR and MAIL). This implies the weighting network can generalize. 

\begin{table}[h!]
\caption{Ablation experiments: $\Delta$, input to the auxiliary network.  Clean test accuracy (Clean), robust test accuracy (PGD) are reported.}
\label{tab:ablat_margin}
\centering
\begin{tabular}{|l|l|l|}
\hline
\multirow{2}{*}{Network input} &  \multicolumn{2}{c|}{CIFAR10} \\ \cline{2-3} 
                  & Clean           & PGD               \\ \hline
$\Delta_i$ (multiclass margin (Def.~\ref{def:mcm}))                &    87.1      &    57.4        \\ \hline \hline
margin (Def.~\ref{def:cm})             &  84.1        &    54.6       \\ 
$\ell(y, f(x + \delta;\theta))$              &  86.9       &  56.9     \\ 
$\ell(y, f(x + \delta;\theta)) - \ell(y, f(x;\theta))$             &  85.4       & 53.8            \\ \hline 
\end{tabular}
\end{table}
In Table~\ref{tab:ablat_margin}, we show that the choice of input to the auxiliary neural network to predict the sample weights has a significant impact. In particular, we show the necessity of using the multi-class margin to achieve superior clean and robust test accuracy. 
Surprisingly, conditioning the weight on the robust loss also leads to good performance, better than the margin , and employing a learnable map for either the class-aware and class-unaware outperforms heuristic methods (e.g., WMMR and MAIL).

\subsection{Ablation study}

In this section, we evaluate variations of our technique on CIFAR-10 using the WRN-32-10 architecture and $\ell_\infty$ with $\epsilon=0.031$. First, we show how the computational cost of \algname{} can be addressed by either utilizing the reweighting network to select a subset of samples on which to do adversarial training or by utilizing alternative attack algorithms that are more efficient compared to PGD. %Next, we evaluate our network in combination with TRADES for various values of $1/\lambda$. 
%
%\input{tables/ablation}
%
\iffalse
\begin{table}[h!]
\caption{Ablation experiments: \algname{}-TRADES with TRADES coefficient $1/\lambda$. Clean test accuracy (Clean), robust test accuracy (PGD) are reported.}
\label{tab:ablat_lambda}
\centering
\begin{tabular}{|l|l|l|}
\hline
\multirow{2}{*}{$1/\lambda$} &  \multicolumn{2}{c|}{CIFAR10} \\ \cline{2-3} 
                  & Clean           & PGD               \\ \hline
$1/\lambda = 6$               &      87.1      &    57.4        \\ \hline \hline
$1/\lambda = 1$            &      87.4    &     52.5      \\ 
$1/\lambda = 5$              &    86.9    &    57.6   \\ 
$1/\lambda = 10$              &    83.8     &  57.9           \\  \hline
\end{tabular}
\end{table}
%
In Table~\ref{tab:ablat_lambda}, we show relative robustness of our approach combined with TRADES to the choice of $1/\lambda$. In particular, we maintain an important advantage of TRADES: the ability to easily control the robustness tradeoff by controlling $1/\lambda$.
\fi

\begin{table}[h!]
\caption{Ablation experiments: Substitution of PGD with F-FGSM~\cite{Wong2020Fast} for the training reweighting steps. Clean test accuracy (Clean), robust test accuracy (PGD) and speedup in train-time over\algname{} are provided.}
\label{tab:ffgsm}
\centering
\begin{tabular}{|l|l|l|l|l|}
\hline
\multicolumn{2}{|c|}{Computation of adv. samples} &  \multicolumn{3}{c|}{CIFAR10} \\ \cline{1-5} 
      Train-step & Reweighting-step            & Clean           & PGD  & Speedup             \\ \hline
PGD & PGD             &    87.1       &    57.4   & $1\times$ 	  \\ 
PGD (80\%) & PGD             &    88.6       &    57.2   & $1.3\times$ 	  \\ 
PGD & F-FGSM           &    88.5      &   57.1   &  $2.6\times$   \\ 
F-FGSM & PGD           &    89.9      &  56.3     &  $4.3\times$  \\ 
F-FGSM & F-FGSM             &  90.1      &    56.1   & $5.8\times$ \\  \hline
\end{tabular}
\end{table}

In the main text, we demonstrate that a pre-trained reweighting network may be used to improve the computational cost of training. In Table~\ref{tab:ffgsm}, we provide ablation experiments on the method used to compute adversarial training and validation samples. Note that the main cost of our algorithm is the computation of adversarial examples to update the classifier and reweighting network. We explore replacing iterative methods (i.e. PGD) with the one-step Fast-FGSM method introduced in \citet{Wong2020Fast}. As a baseline, we explore utilizing the learned weights to \emph{reduce} the computational cost of adversarial training\textemdash i.e. select a subset of each batch to do adversarial training, inspired by~\citet{zhang2020fat}. In the second row of Table~\ref{tab:ffgsm}, we identify 20\% of samples per-batch with the smallest weight. On these samples, we assign $w_i = 0$ (i.e. we perform regular, non-adversarial training). On the rest of the samples, we re-normalize the weights and train as normal using the weighted TRADES loss (computing adversarial perturbations). We see an improvement in runtime and clean test accuracy, and a minor degradation in robust test accuracy. We also explore different combinations of PGD and Fast-FGSM used in the context of \algname{}. For example, we may use PGD to train the classifier (Steps 1 and 3) while using Fast-FGSM to update the weighting network (step 2. Alternatively we could use Fast-FGSM for both. As expected, large improvements in runtime are seen when Fast-FGSM is used (up to $6\times$ when F-FGSM is adopted for both the train and re-weighting step). In other words, it takes 4.3 days (104.2 hours) to train \algname{} using PGD. Using F-FGSM instead of PGD to train the reweighting network results in a reduction in training time to 2 days or 40.1 hours. Gains are largest when F-FGSM is used exclusively. Interestingly, we see only a minor degradation in robust (PGD-based) test accuracy while \emph{improvements} in clean test accuracy are observed.

\begin{table}[h!]
\caption{Ablation experiments:  \algname{}-TRADES with TRADES coefficient $1/\lambda$. Clean test accuracy (Clean), robust test accuracy (PGD) and {AA robust test accuracy} are reported.}
\label{tab:trades_vanilla}
\centering
\begin{tabular}{|l|l|l|l|}
\hline
\multirow{2}{*}{$1/\lambda$} &  \multicolumn{3}{c|}{CIFAR10} \\ \cline{2-4} 
                  & Clean           & PGD  & AA             \\ \hline
BiLAW-TRADES $1/\lambda = 6$               &      87.1      &    57.4   &  	53.6    \\ \hline \hline
TRADES $1/\lambda = 1$            &      87.4    &     52.5  &  45.5   \\ 
TRADES $1/\lambda = 5$              &    86.9    &    57.6   & 52.0\\  \hline
\end{tabular}
\end{table}
We also highlight the relative performance of vanilla TRADES in Table~\ref{tab:trades_vanilla}. When TRADES ($1/\lambda = 1$) and BiLAW with TRADES exhibit similar clean test accuracy, we considerably outperform TRADES with respect to test-set robustness to both PGD and AA-based attacks. When TRADES ($1/\lambda = 6$) and BiLAW exhibit similar AA robustness, we outperform TRADES with respect to clean test-set accuracy and PGD-based robustness. %In particular, we maintain an important advantage of TRADES: the ability to easily control the robustness tradeoff by controlling $1/\lambda$.

\subsection{CIFAR-10 example weights}
\begin{figure}
\centering
\begin{subfigure}[b]{0.47\linewidth}
\includegraphics[width=\linewidth]{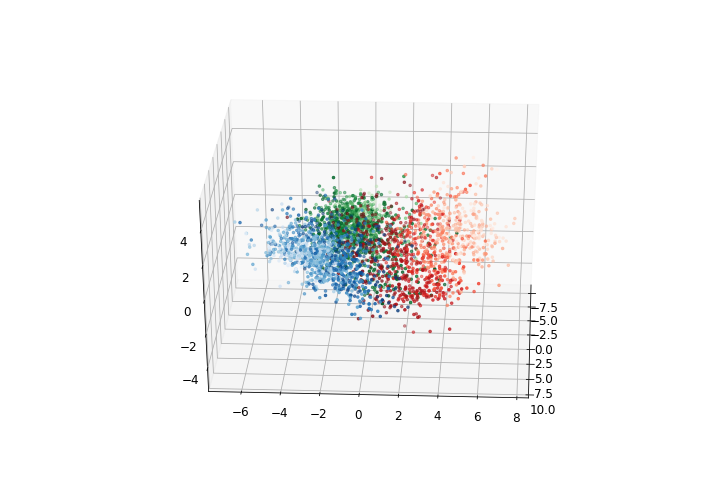}
\caption{}
\end{subfigure}
\begin{subfigure}[b]{0.47\linewidth}
\includegraphics[width=\linewidth]{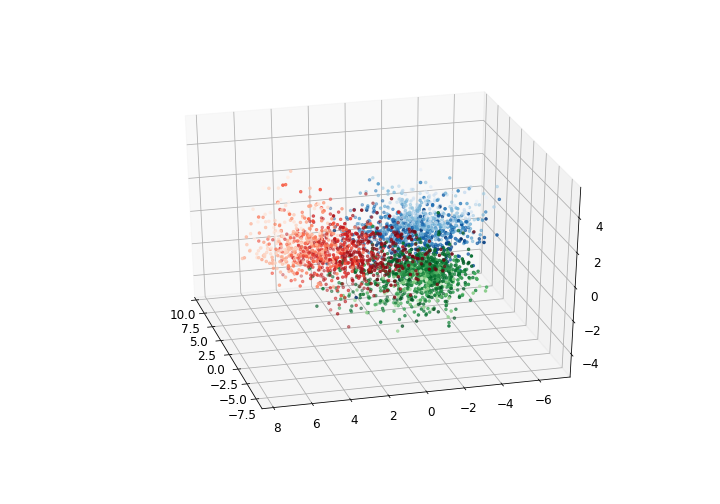}
\caption{}
\end{subfigure}
\caption{Two orientations of a 3-d plot of PCA applied to the model’s likelihood predictions on training samples of three classes from the CIFAR-10 dataset (blue: car, red: plane, \& green: ship). The weight of individual samples (denoted by the shade) correlates with the margin/degree of robustness.}
\label{fig:plane_ship_car}
\end{figure}
\begin{figure}
\centering
\begin{subfigure}[b]{0.47\linewidth}
\includegraphics[trim={0 0 0 0.8cm},clip,width=\linewidth]{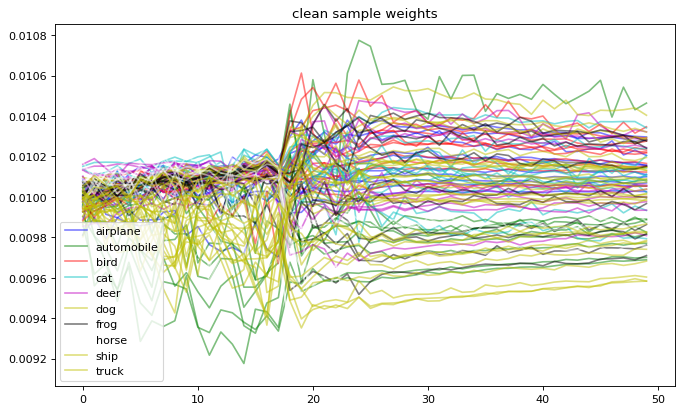}
\caption{}
\end{subfigure}
\begin{subfigure}[b]{0.47\linewidth}
\includegraphics[trim={0 0 0 0.8cm},clip,width=\linewidth]{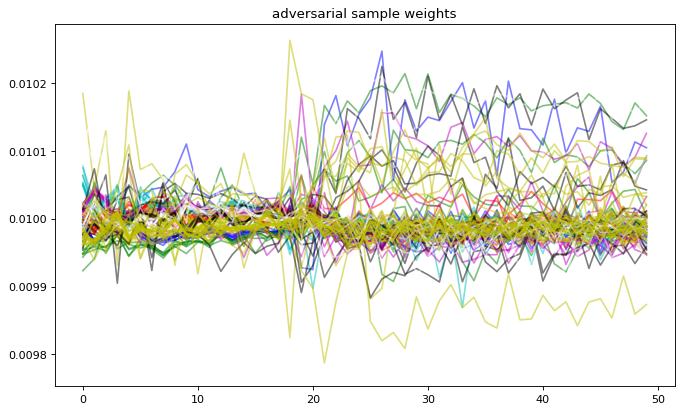}
\caption{}
\end{subfigure}
\caption{(\textbf{a}) Progression of weights associated with a subset of clean training samples (\textbf{b}) Progression of weights associated with a subset of adversarially perturbed training samples}
\label{fig:weight_progression}
\end{figure}
%
%
%\iffalse
\begin{figure}
\centering
\begin{subfigure}[b]{0.34\linewidth}
\includegraphics[width=\linewidth]{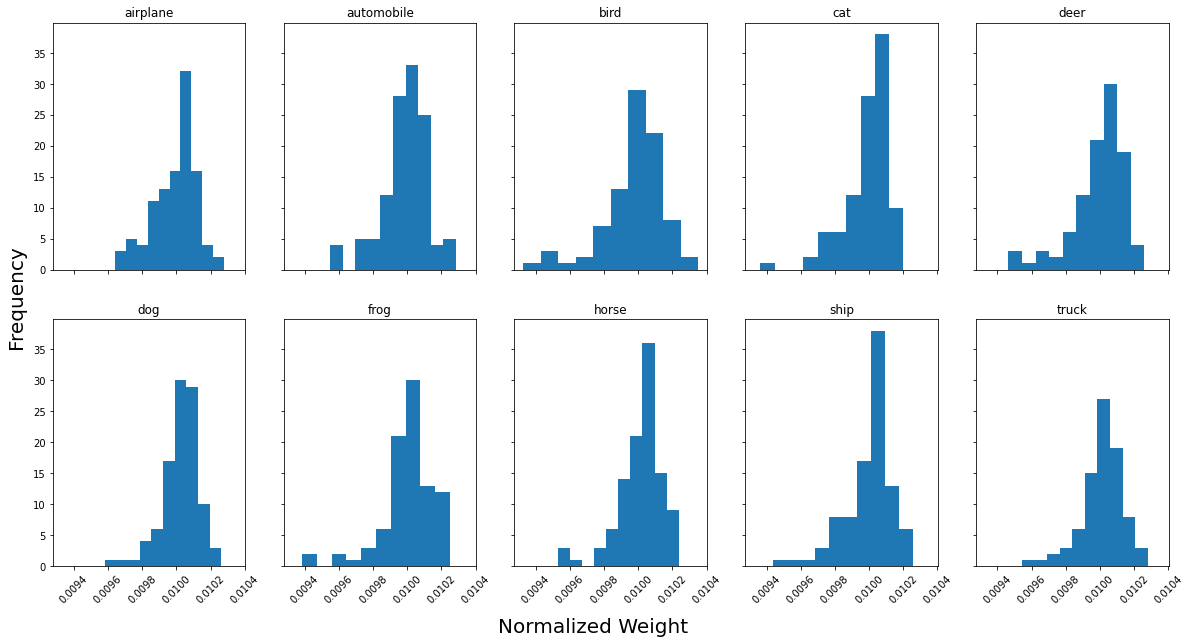}
\caption{}
\end{subfigure}
\begin{subfigure}[b]{0.34\linewidth}
\includegraphics[width=\linewidth]{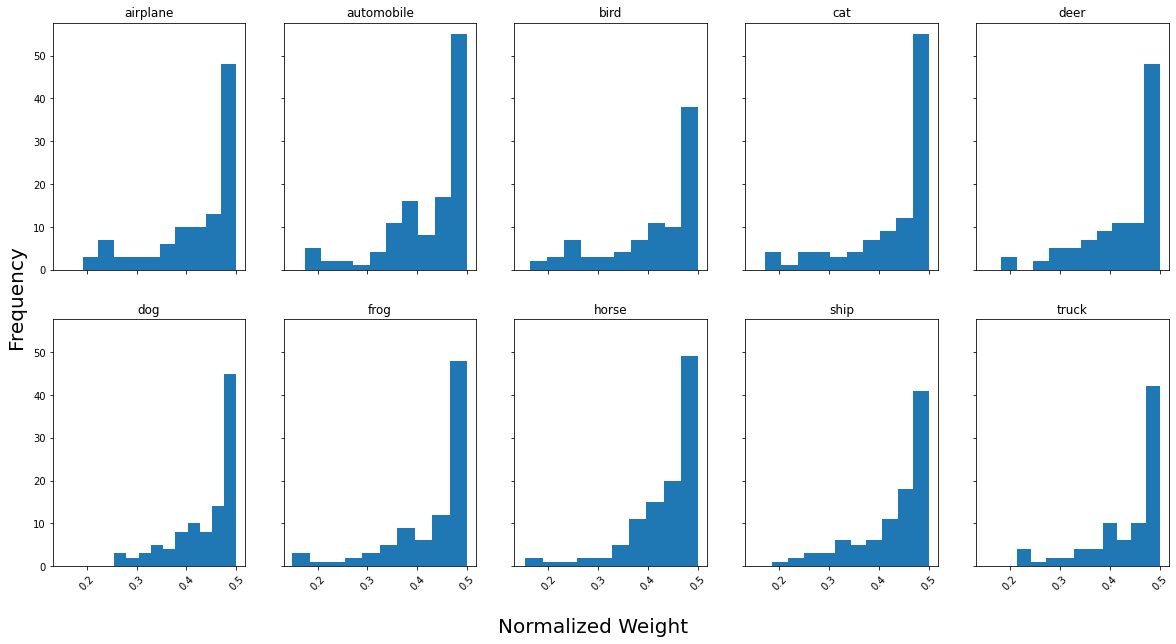}
\caption{}
\end{subfigure}
\begin{subfigure}[b]{0.28\linewidth}
\includegraphics[width=\linewidth]{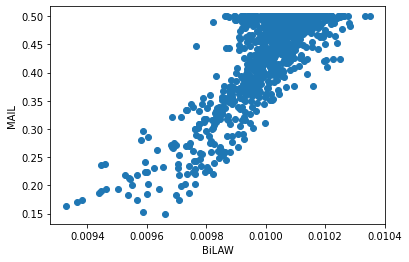}
\caption{}
\end{subfigure}
\begin{subfigure}[b]{0.34\linewidth}
\includegraphics[width=\linewidth]{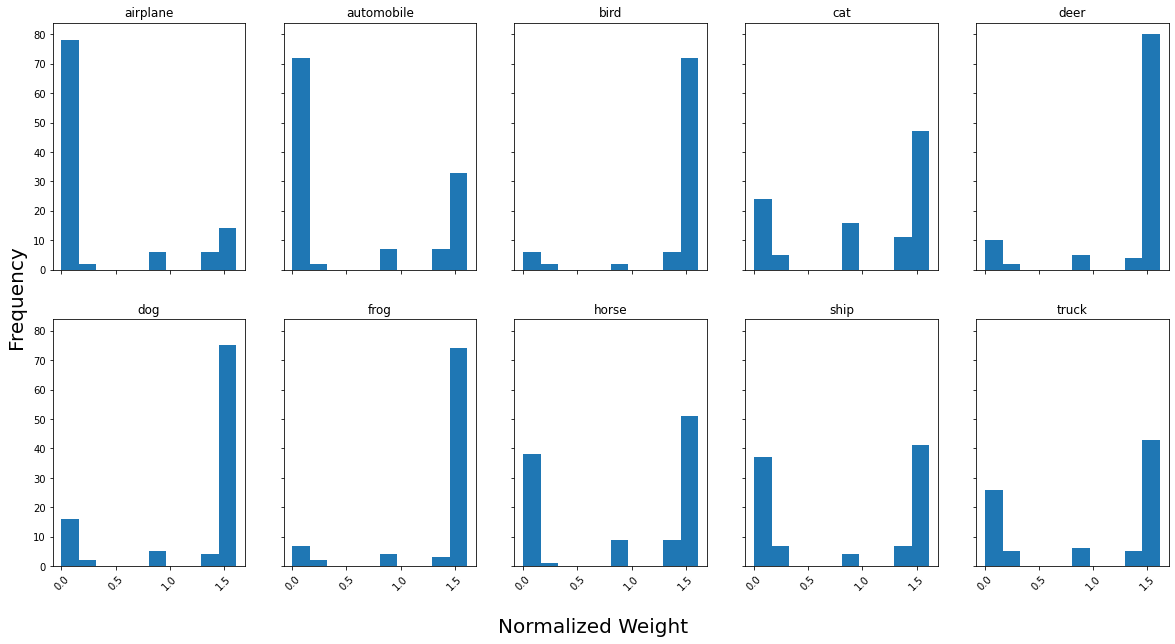}
\caption{}
\end{subfigure}
\begin{subfigure}[b]{0.28\linewidth}
\includegraphics[width=\linewidth]{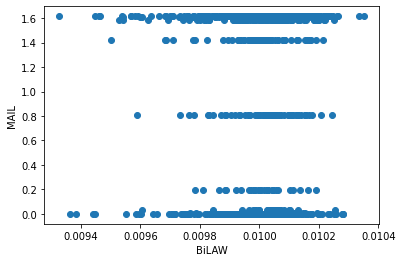}
\caption{}
\end{subfigure}
\caption{(\textbf{a}) BiLAW weight distributions per-class for CIFAR-10 samples. (\textbf{b}) MAIL weight distributions per-class for CIFAR-10 samples.  (\textbf{c}) Scatter plot of MAIL weight vs. BiLAW weight for a robust network. (\textbf{b}) GAIRAT weight distributions per-class for CIFAR-10 samples.  (\textbf{c}) Scatter plot of GAIRAT weight vs. BiLAW weight for a robust network. }
\label{fig:mail_bilaw_weights}
\end{figure}
%\fi
%
In Fig~\ref{fig:plane_ship_car} we recover the predictions made by a small-CNN trained with \algname{}. We then use principal component analysis (PCA) to project 10-dimensional predicted class likelihoods into 2-dimensions and plot the corresponding embeddings. The color denotes the degree of the robustness of each data point. Samples which are assigned larger weight are darker. As expected, these samples associated with high weights lie close to the decision boundary and are more likely to improve robust generalization.

In Fig~\ref{fig:weight_progression} we investigate the dynamics of predicted weights by visualizing the progression of weights predicted at margins for training samples and their adversarial variants. We observe (1) the dynamics of the weights seem to be determined largely by the learning rate of the classifier (i.e. the first adjustment to the learning rate happens around epoch 20), (2) the majority of weights predicted for \textit{clean samples} are low (i.e. most clean samples are easy), and (3) the variance of the weight distribution is quite tight for adversarial samples.

In Fig~\ref{fig:mail_bilaw_weights} we compare weights computed via the GAIRAT and MAIL heuristics to weights predicted via BiLAW and show a positive correlation. In particular, BiLAW may be considered a generalization of the MAIL heuristic that additionally incorporates multi-class margin information. The similarity between margin-based weight estimators BiLAW and MAIL is evident, while the PGD-based GAIRAT weighting heuristic emphasizes a bimodal weight distribution.

Replicating (Fig.~\ref{fig:cifarexamples}), we plot samples with small and large weight for competitive methods GAIRAT and MAIL. As with out method, samples associated with small weights appear to be ``easy'' and visa versa.
\begin{figure}[h]
    \centering
    \includegraphics[width=\textwidth]{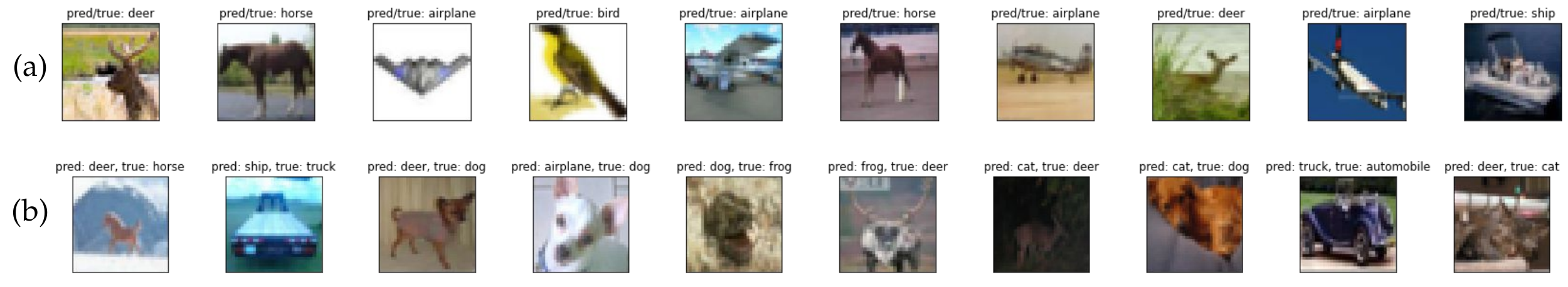}
    \caption{Examples taken from CIFAR-10 and weighted using GAIRAT~\citep{zhang2021geometryaware}. (\textbf{a}) Samples with low weight. (\textbf{b}) Samples with high weight.}
    \label{fig:mnistexamples}
\end{figure}

\begin{figure}[h]
    \centering
    \includegraphics[width=\textwidth]{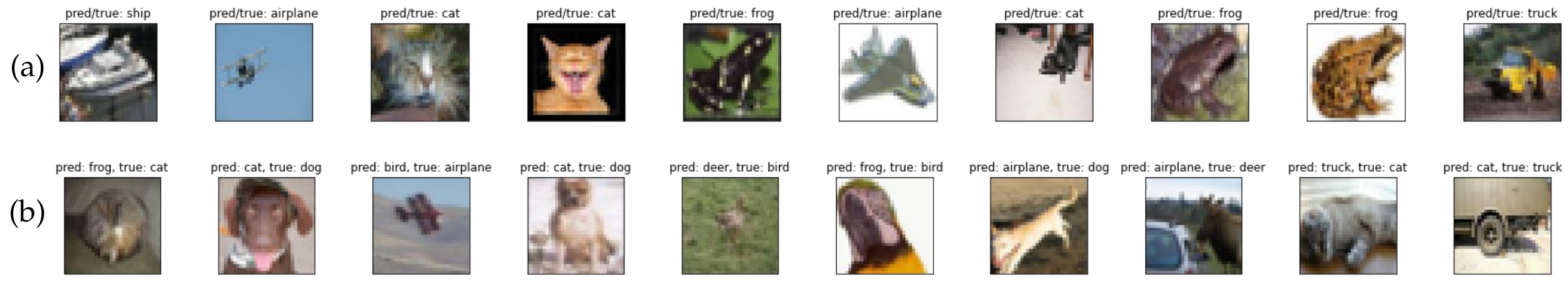}
    \caption{Examples taken from CIFAR-10 and weighted using MAIL~\citep{wang2021probabalisticmarginat}. (\textbf{a}) Samples with low weight. (\textbf{b}) Samples with high weight.}
    \label{fig:mnistexamples}
\end{figure}

\subsection{MNIST Experiments}
\label{app:mnist} 
\begin{table*}[ht]
\caption{MNIST/F-MNIST comparison for plain, AT, GAIRAT, WMMR ($\alpha_{\textrm{train}} = 0.1$, $\alpha_{\textrm{test}} = 2$), MAIL ($\gamma = 5$, $\beta = 0.05$) and \textbf{\algname{}} using standard robust loss. 
Clean test accuracy (Clean), robust test accuracy (PGD) and {AA robust test accuracy} are reported.
\textbf{\underline{Best}} result is underlined and bolded and \textbf{second best} is bolded.} % The best result is marked by an underlined and \textbf{\underline{bold  value}} and the second best by \textbf{bold value}.
\label{tab:results}
\centering
\resizebox{0.8\textwidth}{!}{%
\begin{tabular}{|l| l l l ||l l l ||l l l ||l l l |} 
\hline

\multirow{2}{*}{}    & \multicolumn{6}{c||}{ Tiny-CNN } & \multicolumn{6}{c|}{ FC1 }     \\ 

\cline{2-13}

\multirow{2}{*}{} & \multicolumn{3}{l||}{perturbation: $\ell_\infty$ } & \multicolumn{3}{l||}{perturbation: $\ell_2$ }      & \multicolumn{3}{l||}{perturbation:~$\ell_\infty$ }   & \multicolumn{3}{l|}{perturbation:~$\ell_2$}  \\ 

%\cline{2-9}

       & Clean    & PGD   & {AA}          & Clean    & PGD  & {AA}               & Clean    & PGD    & {AA}            & Clean    & PGD     & {AA}               \\ 
\hline
\hline
\textit{MNIST}       & \multicolumn{3}{c||}{ $\epsilon = 0.1$}  & \multicolumn{3}{c||}{ $\epsilon = 0.3$ } & \multicolumn{3}{c||}{ $\epsilon = 0.1$ }    & \multicolumn{3}{c|}{ $\epsilon = 0.3$ }        \\ 

\hline
plain                               & \textbf{99.1}       & 21.7                &{{9.1}}                 & \underline{ \textbf{99.2}} & 96.9                & {36.4}                & 98.4                & 1.7                 &{{0.0}}                 & 98.3                & 90.3                &   {16.1} \\
AT                                  & 99.0                & \textbf{95.9}       &\underline{ \textbf{{93.7}}} & \textbf{99.1}       & 98.2                & \underline{ \textbf{{96.1}}} & 98.4                & 92.9                &{{90.4}}                & 8.8                 & \textbf{97.4}       &    \underline{ \textbf{{95.3}}}                \\
GAIRAT                              & \textbf{99.1}       & \underline{ \textbf{96.7}} &{{91.1}}                & \underline{ \textbf{99.2}} & \textbf{98.8}       & {90.3}                & \textbf{99.0}       & \underline{ \textbf{93.2}} &{{89.7}}                & \underline{ \textbf{98.8}} & \underline{ \textbf{97.6}} &   {89.2}                   \\
WMMR                                & 98.8                & 94.3                &{{90.2 }}               & 99.0                & 98.5                & {91.7}                & 98.9                & 92.8                &{{89.4}}                & 98.2                & 97.2                &       {89.8}              \\
MAIL                                & 98.6                & 95.1                &{{91.4}}                & 98.7                & 98.6                & \textbf{{95.4}}       & 98.4                & \textbf{93.1}       &{\textbf{{91.3}}}       & 98.1                & \textbf{97.4}       &     {94.2}                \\
\algname(ours) & \underline{ \textbf{99.2}} & \underline{ \textbf{96.7}} &{\textbf{{91.7}}}       & \underline{ \textbf{99.2}} & \underline{ \textbf{98.9}} & \textbf{{95.4}}       & \underline{ \textbf{99.1}} & \textbf{93.1}       &{{\underline{ \textbf{91.6}}}} & \textbf{98.6}       & \underline{ \textbf{97.6}} &    \textbf{{94.4}}                 \\ \hline
\textit{F-MNIST}       & \multicolumn{3}{c||}{ $\epsilon = 0.1$ } & \multicolumn{3}{c||}{ $\epsilon = 0.3$ }    & \multicolumn{3}{c||}{ $\epsilon = 0.1$ } & \multicolumn{3}{c|}{ $\epsilon = 0.3$ }      \\ 
\hline
plain                               & \textbf{89.6}       & 1.5                 & {0.0}                & 89.7                & 42.9                & {0.0}                 & \underline{ \textbf{98.5}} & 0.0                 &{{0.0}}                 & 89.3                & 57.2                & {0.0}                 \\
AT                                  & 86.4                & 70.1                &{{68.3}}                & 91.9                & 79.6                & {\underline{ \textbf{77.9}}} & 87.0                & 68.7                &{{66.3}}                & 89.8                & 80.1                & \textbf{{76.0}}       \\
GAIRAT                              & 86.4                & \underline{ \textbf{77.6}} &{{64.3}}                & \textbf{92.3}       & \textbf{81.1}       & {70.3}                & 87.1                & \textbf{70.2}       &{{61.4}}                & 91.1                & \underline{ \textbf{81.0}} & {70.4}                \\
WMMR                                & 86.2                & 77.3                &{{64.1}}                & 92.1                & 80.6                & {71.4}                & 86.9                & 68.4                &{{61.3}}                & 91.1                & 78.4                & {70.9}                \\
MAIL                                & 86.4                & 76.9                &{\textbf{{68.6}}}       & 92.2                & 80.5                & {76.2}                & 90.1                & 69.3                &{\textbf{{66.4}}}       & 90.6                & 79.3                & {75.9}                \\
\algname(ours)  & \textbf{86.6}       & \textbf{77.4}       &{{\underline{ \textbf{68.8}}}} & \underline{ \textbf{92.4}} & \underline{ \textbf{81.3}} & \textbf{{76.6}}       & 87.3                & \underline{ \textbf{70.6}} &{{\underline{ \textbf{66.7}}}} & \underline{ \textbf{91.4}} & \textbf{80.9}       & {\underline{ \textbf{76.1}}} \\ \hline
\arrayrulecolor{white}
\end{tabular}}
\end{table*}

In Table~\ref{tab:results}, we evaluate \algname{} using two relatively small networks on two datasets: MNIST~\citep{lecun-mnisthandwrittendigit-2010} and Fashion MNIST~\citep{Xiao17}. Tiny-CNN is a convolutional network with 2 convolutional and 2 dense layers. FC1 corresponds to a single hidden layer feedforward network with 1024 hidden units. The details of the architectures are given in Appendix~\ref{app:architecture}. We consider robustness with respect to $\ell_\infty$ distance. We use three criteria: clean test accuracy (clean), robust test accuracy (PGD) for a given threshold $\epsilon$ and AutoAttack (AA). Robust test accuracy is computed using Projected Gradient Descent (PGD)~\citep{madry2018towards} with 20 iterations. In all testcases, our method matches the performance of GAIRAT and out-performs the other methods for  clean and PGD accuracy and we out-perform all reweighting methods on AA accuracy. However, we note the overall distribution of both clean and robust accuracy is tight. %and potentially an artifact of how we set various hyperparameters. 
We note a potential drawback of reweighting algorithms: the MNIST and F-MNIST datasets contain a non-trivial number of misclassified samples which can influence performance~\citep{mislabeled}. For algorithms which perform weighted training, possible large weights on outliers or mislabeled examples may influence classification performance. We will investigate this in the context of adversarial training in future work.

We plot MNIST samples with small and large weight. As with CIFAR-10 (Fig.~\ref{fig:cifarexamples}), samples associated with small weights appear to be ``easy'' in the sense that the digits are neatly written. On the other hand, digits associate with high weight are easily confused and often involve the occurrence or lack of occurrence of spaces between strokes that define certain digits (e.g. 3, 5, 0,  9, and 8).
\begin{figure}[h]
    \centering
    \includegraphics[width=\textwidth]{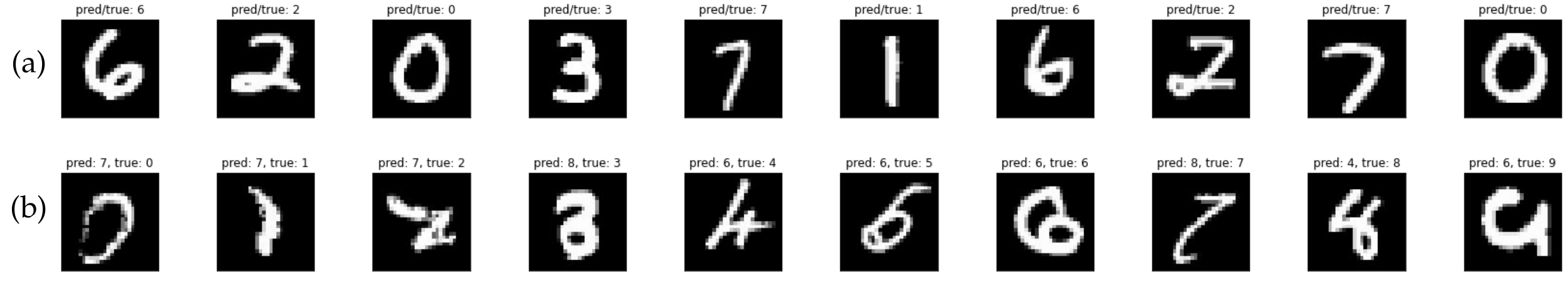}
    \caption{Examples taken from MNIST. (\textbf{a}) Samples with low weight. (\textbf{b}) Samples with high weight.}
    \label{fig:mnistexamples}
\end{figure}

\end{document}